\documentclass[10pt,twocolumn,letterpaper]{article}
\usepackage{ijcb}
\usepackage{booktabs}
\usepackage{times}
\usepackage{epsfig}
\usepackage{graphicx}
\usepackage{amsmath}
\usepackage{amssymb}
\usepackage{url}
\usepackage{caption}
\usepackage{subcaption}
\usepackage{float}
\usepackage{makecell}
\usepackage{enumitem}
\usepackage{pifont}
\usepackage{hyperref}
\usepackage{array}
\usepackage{fixltx2e}
\usepackage{appendix}
\usepackage{lipsum}
\usepackage{fancyhdr}
\newcolumntype{C}[1]{>{\centering\arraybackslash}p{#1}}
\newcolumntype{P}[1]{>{\centering\arraybackslash}m{#1}}
\usepackage[flushleft]{threeparttable}
\ijcbfinalcopy 

\ifijcbfinal\fi
\begin{document}

\fancypagestyle{title}{%
  \setlength{\headheight}{10pt}%
  \fancyhf{}
  \renewcommand{\headrulewidth}{0pt}
  \fancyhead[C]{\Large{\textcolor{red}{To appear in International Joint Conference on Biometrics (IJCB), 2020}}}
}%

\title{Fingerprint Synthesis: Search with 100 Million Prints}

\author{Vishesh Mistry\\
Michigan State University\\
East Lansing, MI, USA\\
{\tt\small mistryvi@msu.edu}
\and
Joshua J. Engelsma\\
Michigan State University\\
East Lansing, MI, USA\\
{\tt\small engelsm7@cse.msu.edu}
\and
Anil K. Jain\\
Michigan State University\\
East Lansing, MI, USA\\
{\tt\small jain@cse.msu.edu}
}

\twocolumn[{%
\renewcommand\twocolumn[1][]{#1}%
\maketitle
 \thispagestyle{title}
\begin{center}
\vspace{-1.2em}
    \centering
    \footnotesize
    \captionsetup{font=footnotesize}
    \setlength{\fboxsep}{2.2pt}
    \setlength{\fboxrule}{0.2pt}
    \begin{minipage}{\linewidth}
    \fcolorbox{white}{white}{\includegraphics[width=0.24\linewidth]{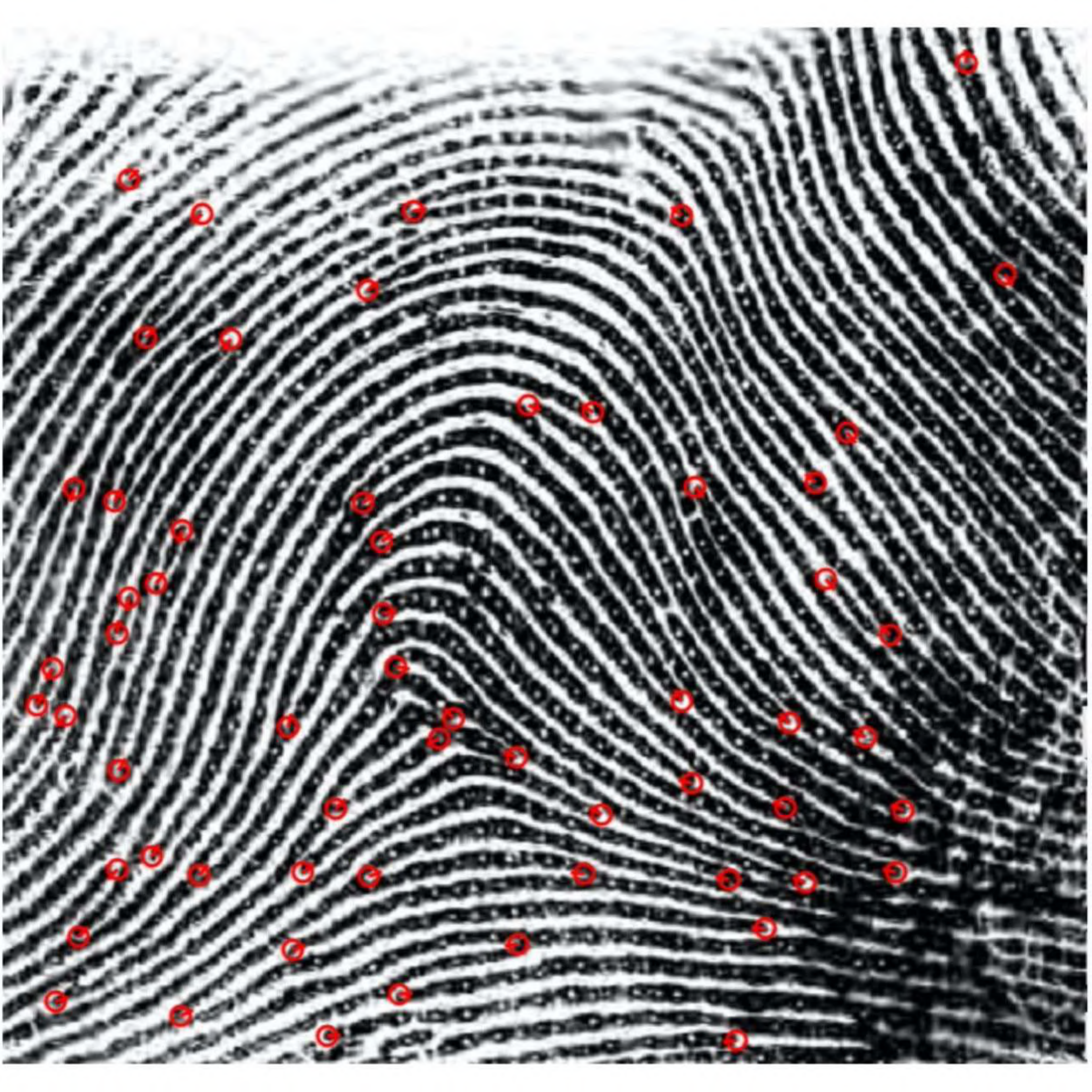}}\hfill
    \fcolorbox{white}{white}{\includegraphics[width=0.24\linewidth]{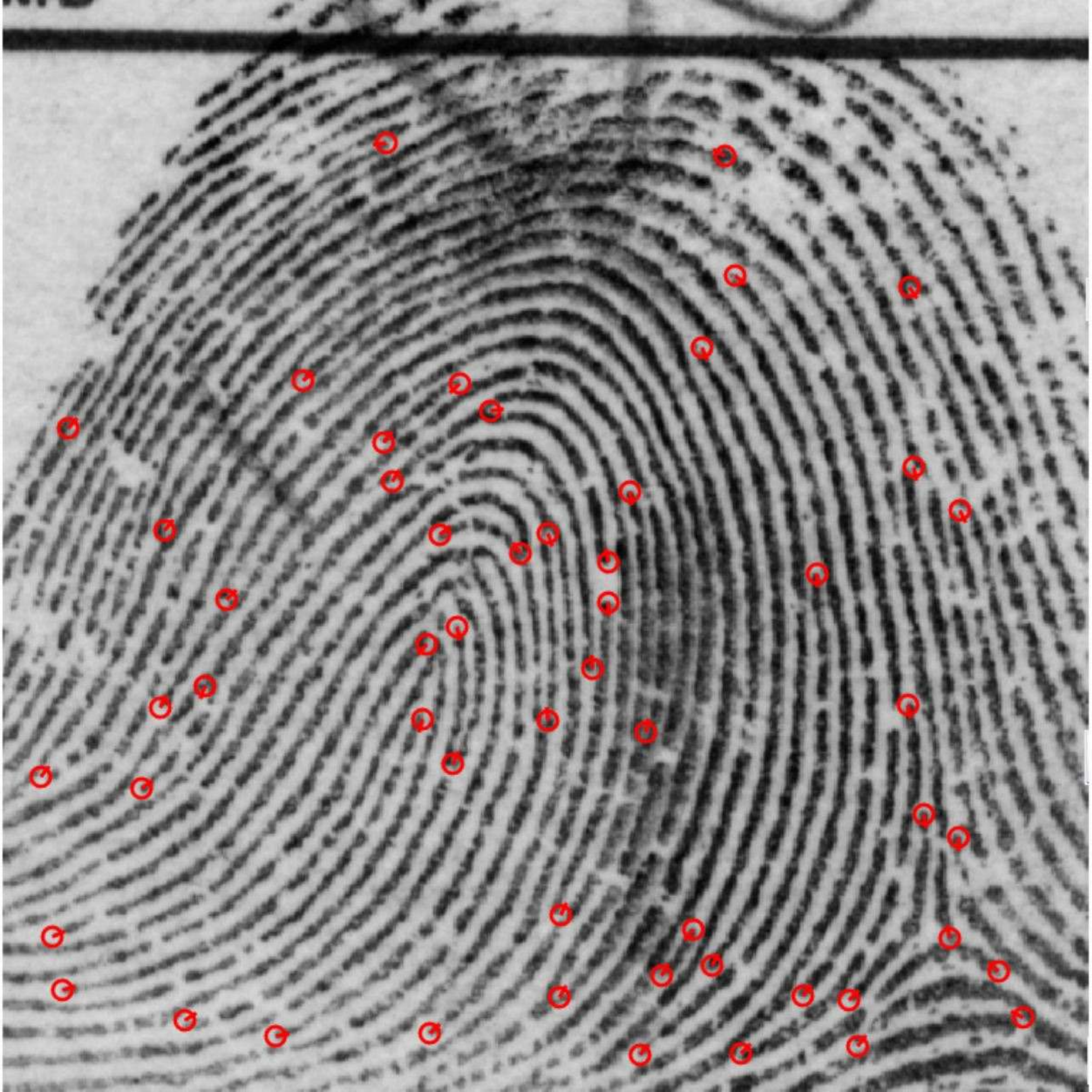}}\hfill
    \fcolorbox{white}{white}{\includegraphics[width=0.24\linewidth]{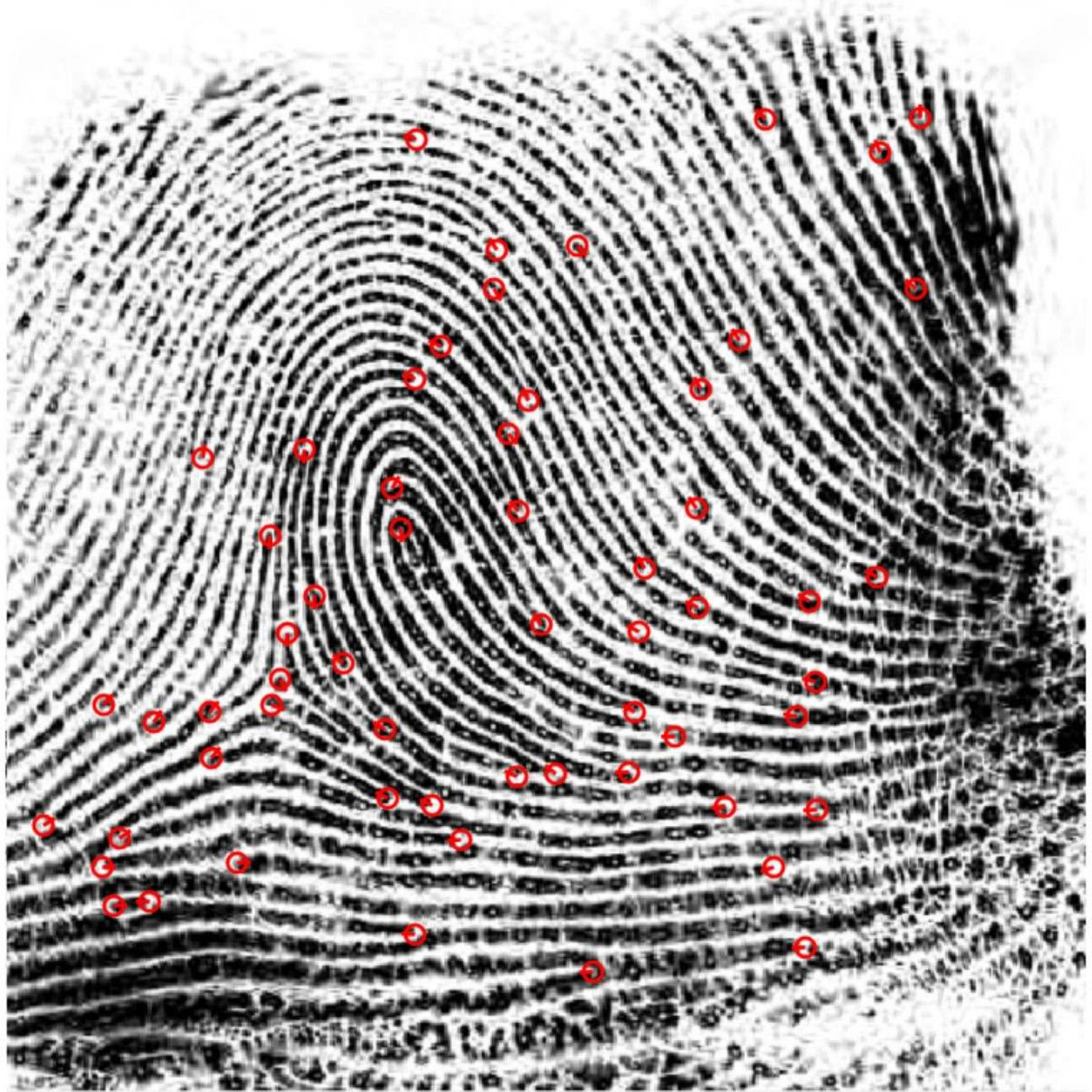}}\hfill
    \fcolorbox{white}{white}{\includegraphics[width=0.24\linewidth]{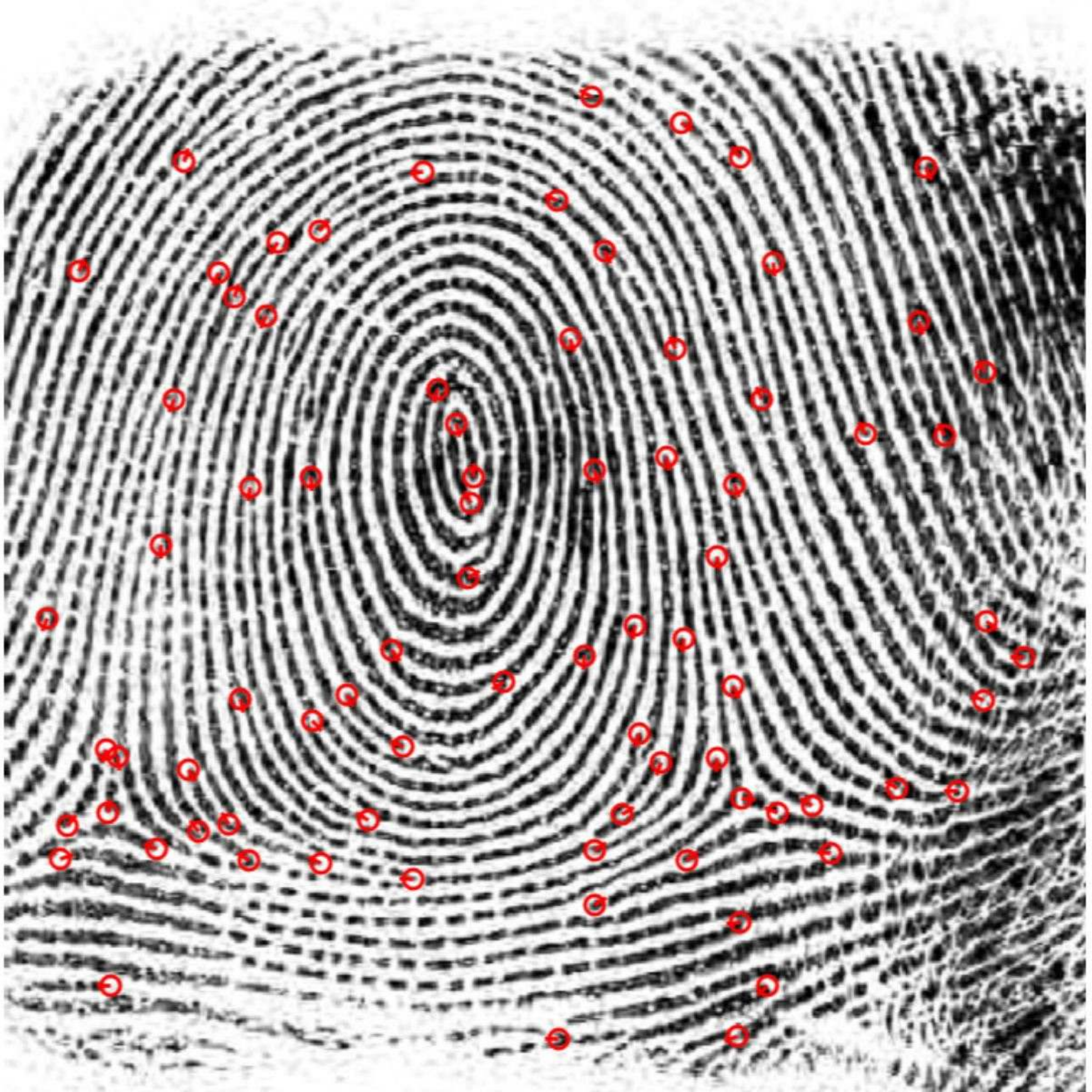}}\hfill
    \centering {\small(a)}
    \end{minipage}\\
    \begin{minipage}{\linewidth}
    \fcolorbox{white}{white}{\includegraphics[width=0.24\linewidth]{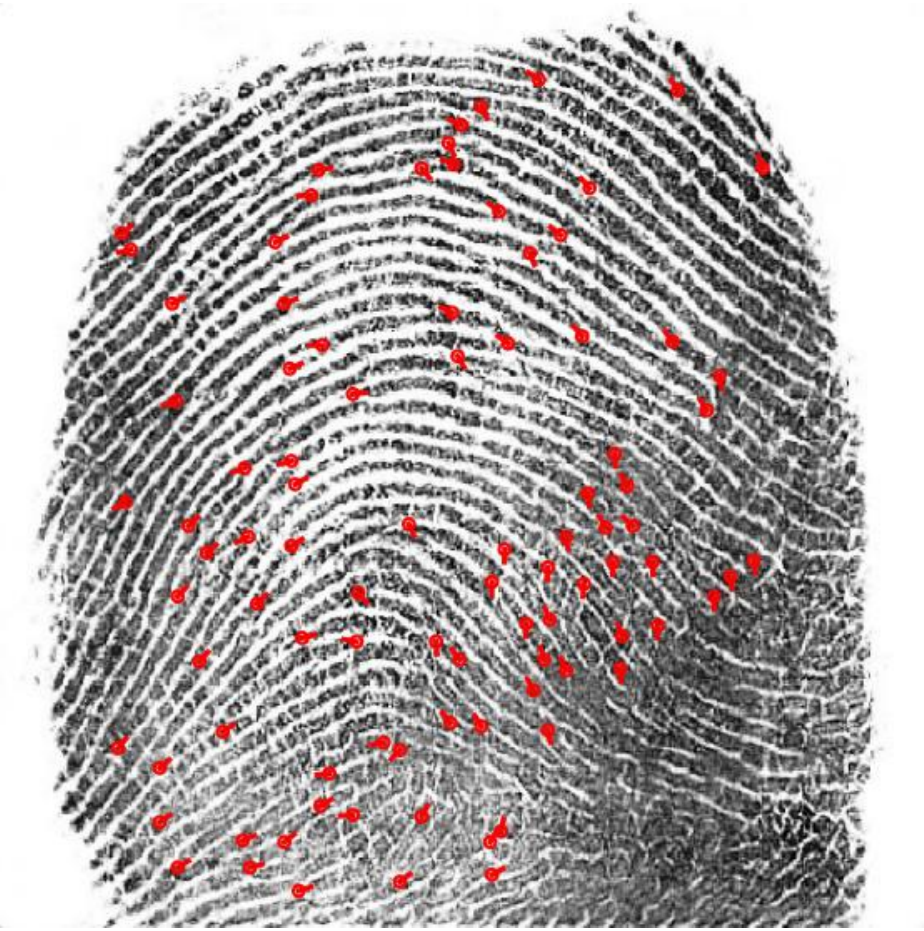}}\hfill
    \fcolorbox{white}{white}{\includegraphics[width=0.24\linewidth]{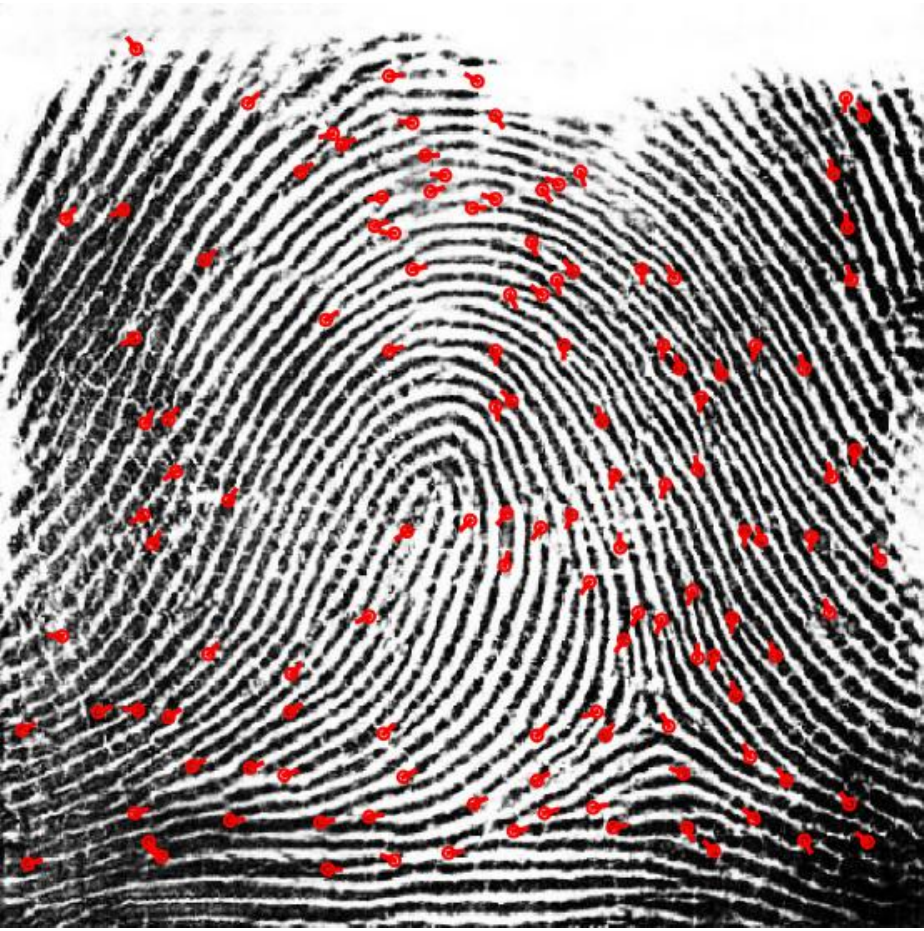}}\hfill
    \fcolorbox{white}{white}{\includegraphics[width=0.24\linewidth]{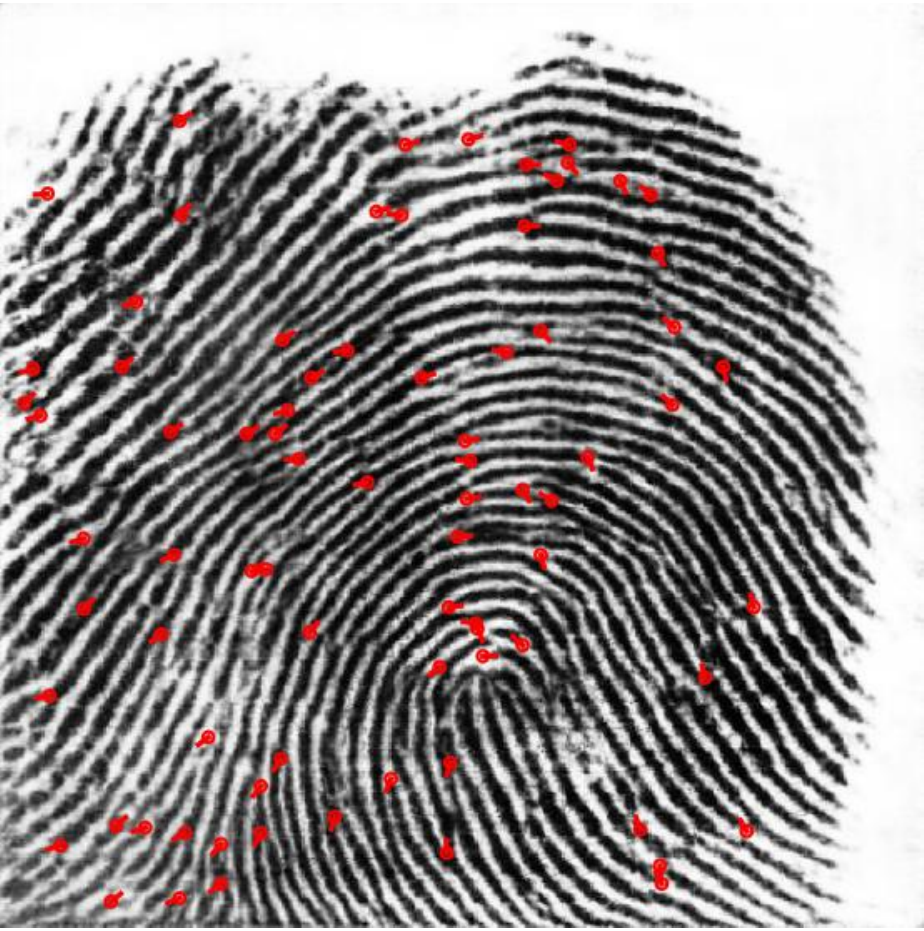}}\hfill
     \fcolorbox{white}{white}{\includegraphics[width=0.24\linewidth]{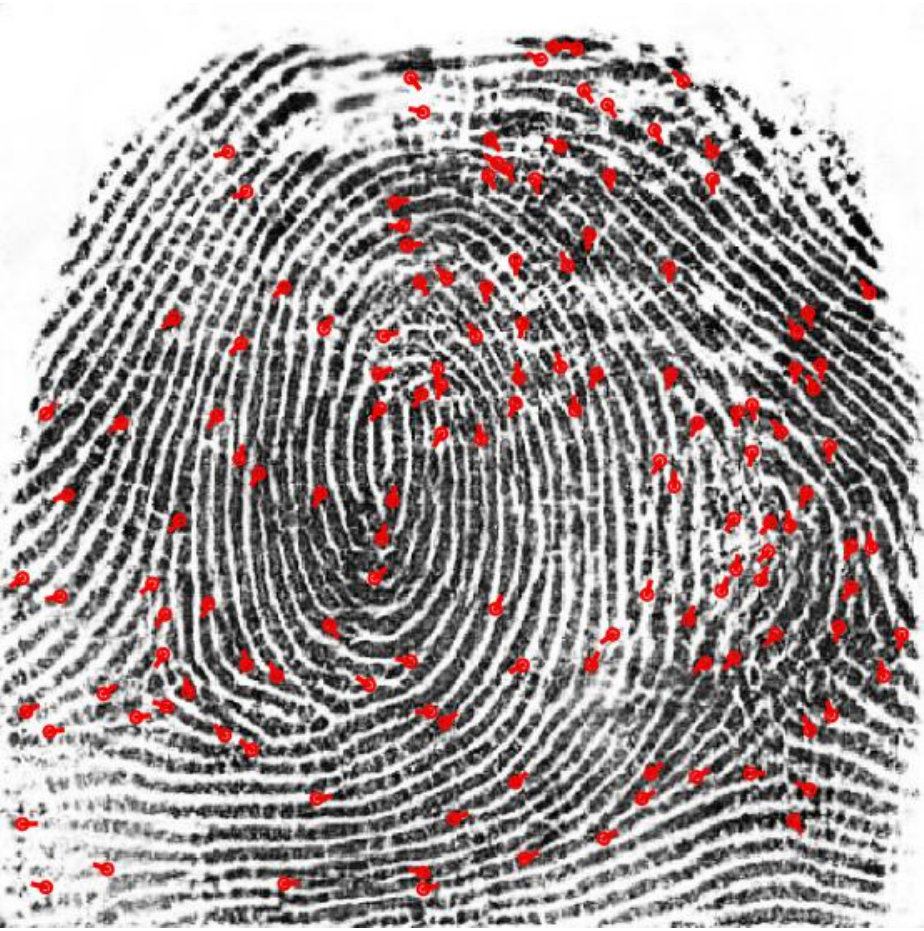}}\hfill
    \centering {\small(b)}
    \end{minipage}\;
    \captionof{figure}{Comparison of real fingerprints with synthesized fingerprints. Images in (a) are rolled fingerprint images from a law enforcement operational dataset \cite{kai_fing} and fingerprints in (b) are synthesized using the proposed approach. One example for each of the four major fingerprint types is shown (from left to right: arch, left loop, right loop, whorl). All the fingerprints have dimensions of $512\times512$ pixels and a resolution of 500 dpi. A COTS minutiae extractor was used to extract the minutiae points overlaid on the fingerprint images. Fingerprints in (a), from left to right, have NFIQ 2.0 quality scores of (47, 64, 77, 73), respectively, while those in (b), from left to right, have quality scores of (55, 65, 60, 69), respectively.}
    \label{fig:real_vs_extracted}
\end{center}
}]

\begin{abstract}
   Evaluation of large-scale fingerprint search algorithms has been limited due to lack of publicly available datasets. To address this problem, we utilize a Generative Adversarial Network (GAN) to synthesize a fingerprint dataset consisting of 100 million fingerprint images. In contrast to existing fingerprint synthesis algorithms, we incorporate an identity loss which guides the generator to synthesize fingerprints corresponding to more distinct identities. The characteristics of our synthesized fingerprints are shown to be more similar to real fingerprints than existing methods via eight different metrics (minutiae count - block and template, minutiae direction - block and template, minutiae convex hull area, minutiae spatial distribution, block minutiae quality distribution, and NFIQ 2.0 scores). Additionally, the synthetic fingerprints based on our approach are shown to be more distinct than synthetic fingerprints based on published methods through search results and imposter distribution statistics. Finally, we report for the first time in open literature, search accuracy against a gallery of \textbf{100 million fingerprint images}  (NIST SD4 Rank-1 accuracy of 89.7\%).
\end{abstract}

\begin{table*}[t]
\small
    \centering
    \begin{threeparttable}
    \begin{tabular}{ P{2.1cm}P{2.1cm}P{2.2cm}P{2.4cm}P{7.0cm} }
    \toprule
    \textbf{Algorithm} & \textbf{Orientation Model} & \textbf{Ridge Structure Model} & \textbf{Minutiae Model} & \textbf{Analysis}\tnote{1}\\
    \midrule
    Cappelli \textit{et al.} \cite{cappeli} & Zero-pole \cite{zero_pole} & Gabor Filter~\cite{gabor} & Gabor Filter~\cite{gabor} & \begin{itemize}[nolistsep, before*={\mbox{}\vspace{-\baselineskip}}, after*={\mbox{}\vspace{-\baselineskip}}]
        \item Generated 10K plain fingerprints ($240\times336$)
    \end{itemize}{}\\
    \midrule
    {SFinGe}~\cite{sfinge,cappelli_iet} & Zero-pole~\cite{zero_pole} & Gabor Filter~\cite{gabor} & Gabor Filter~\cite{gabor} & 
    \begin{itemize}[nolistsep, before*={\mbox{}\vspace{-\baselineskip}}, after*={\mbox{}\vspace{-\baselineskip}}]
        \item Generated 100K plain-prints per 24 hours
    \end{itemize}{}\\
    \midrule
    Johnson \textit{et al.} \cite{johnson} & Zero-pole \cite{zero_pole} & Gabor Filter~\cite{gabor} & Spatial~\cite{spatial_model} & \begin{itemize}[nolistsep, before*={\mbox{}\vspace{-\baselineskip}}, after*={\mbox{}\vspace{-\baselineskip}}]
        \item Generated 2,400 plain-prints
    \end{itemize}{}\\
    \midrule
    Zhao \textit{et al.}~\cite{zhao} & Zero-pole \cite{zero_pole} & AM-FM~\cite{am_fm} & Spatial~\cite{spatial_model} & \begin{itemize}[nolistsep, before*={\mbox{}\vspace{-\baselineskip}}, after*={\mbox{}\vspace{-\baselineskip}}]
        \item Generated 5,000 rolled-prints.
    \end{itemize}{}\\
    \midrule
    Bontrager \textit{et al.} \cite{fing_gan} & \multicolumn{3}{P{6.7cm}}{Wasserstein GAN (WGAN) \cite{wgan}} & \begin{itemize}[nolistsep, before*={\mbox{}\vspace{-\baselineskip}}, after*={\mbox{}\vspace{-\baselineskip}}]
        \item Generated $128\times128$ fingerprint patches
    \end{itemize}{}\\
    \midrule
    Cao \& Jain \cite{kai_10mil} & \multicolumn{3}{P{6.7cm}}{IWGAN \cite{iwgan} and Autoencoder~\cite{cae}} & \begin{itemize}[nolistsep, before*={\mbox{}\vspace{-\baselineskip}}, after*={\mbox{}\vspace{-\baselineskip}}]
        \item Synthesized 10 million rolled-prints
        \item Fingerprints lack ``uniqueness"
    \end{itemize}{}\\
    \midrule
    Attia \textit{et al.}~\cite{attia} & \multicolumn{3}{P{6.7cm}}{Variational Autoencoder~\cite{vae_1,vae_2}} & \begin{itemize}[nolistsep, before*={\mbox{}\vspace{-\baselineskip}}, after*={\mbox{}\vspace{-\baselineskip}}]
        \item Insufficient training data (800 unique fingerprints)
        \item No evaluation of ``uniqueness"
    \end{itemize}{}\\
    \midrule
    \textbf{Proposed Approach} & \multicolumn{3}{P{6.7cm}}{IWGAN \cite{iwgan} and Autoencoder~\cite{cae} with Identity Loss~\cite{deepprint}} & \begin{itemize}[nolistsep, before*={\mbox{}\vspace{-\baselineskip}}, after*={\mbox{}\vspace{-\baselineskip}}]
        \item Synthesized \& searched \textbf{100 million} fingerprints
        \item Improved realism and uniqueness
    \end{itemize}{}\\
    \bottomrule
    \end{tabular}
    \begin{tablenotes}
    \item[1] Generally speaking, model-based approaches lack ``realism" in comparison to the more recent learning-based approaches.
    \end{tablenotes}
    \end{threeparttable}
    \caption{Summary of Fingerprint Synthesis Methods in the Literature}
    \label{tab:synthesis_literature}
    \vspace{-1.0em}
\end{table*}{}

\section{Introduction}

Large-scale Automated Fingerprint Identification Systems (AFIS) have been widely deployed throughout law enforcement and forensics agencies, international border crossings, and national ID systems. A few notable applications, where large-scale AFIS are being used include: (i) India's Aadhaar Project~\cite{aadhaar} (1.25 billion ten-prints), (ii) The FBI's Next Generation Identification system (NGI)~\cite{ngi} (145.3 million ten-prints), and (iii) The DHS's Office of Biometric Identity Management (OBIM) system~\cite{obim} (fingerprints and face images of non-US citizens at ports of entry to the United States).

While these large-scale search systems are seemingly quite successful, little work has been done in the academic literature to understand the performance of these search systems at scale (search performance on a gallery of 100 million or larger). In particular, a number of fingerprint search algorithms have been proposed~\cite{fing_index_1,kai_fing, fing_index_3, fing_index_4, fing_index_5}, but nearly all of them were evaluated on a gallery of at most 2,700 rolled fingerprints from NIST SD14~\cite{nist_sd14}. A few exceptions include~\cite{fing_index_1, deepprint} and~\cite{kai_fing} with gallery sizes of 1 million and 250,000, respectively. Given this limitation in our understanding of large-scale search performance, a primary contribution of this paper is the evaluation of fingerprint search performance (accuracy and efficiency) against a gallery at a scale of 100 million fingerprints.

To evaluate against a gallery of 100 million requires one to either (i) collect large-scale fingerprint records with Institutional Review Board (IRB) approval, (ii) obtain 100 million fingerprint records from an existing forensic or government dataset, or (iii) synthesize fingerprint images. 

Collecting 100 million fingerprints is practically infeasible in terms of cost, time, and IRB regulations. Furthermore, governmental privacy regulations\footnote{\href{https://bit.ly/2YD4e4A}{https://bit.ly/2YD4e4A}} prohibit sharing of existing fingerprint datasets (say from National ID systems). Indeed, NIST has even taken down some of their previously public fingerprint datasets (NIST SD27~\cite{nist_sd27}, NIST SD14~\cite{nist_sd14}, and NIST SD4~\cite{nist_sd4}) due to stringent privacy regulations. Therefore, to adequately evaluate large-scale fingerprint search algorithms, we propose to first synthesize 100 million fingerprints with characteristics close to those of real fingerprint images and later, to use our synthetic fingerprints to augment the gallery for search performance evaluation at scale.

More concisely, the contributions of this research are:

\begin{itemize}
    \item  A fingerprint synthesis algorithm built upon GANs~\cite{gan} which is trained on a large dataset of real fingerprints~\cite{msp_long}. We show that our \textit{learning-based} synthesis approach generates more realistic fingerprints (both plain-prints and rolled-prints) than traditional \textit{model-based} synthesis algorithms~\cite{zhao,cappeli,sfinge,johnson}.
    \item In contrast to existing learning-based synthesis algorithms~\cite{kai_10mil,attia,fing_gan,finger_gan}, we add an identity loss to generate fingerprints of more \textit{unique} identities.
    \item Synthesis of 100 million fingerprint images. This synthesis required code optimization and resources (51 hours of run time on 100 Tesla K80 GPUs).
    
    \twocolumn[{
\renewcommand\twocolumn[1][]{#1}
\begin{center}
    \centering
    \footnotesize
    \captionsetup{font=footnotesize}
    \setlength{\fboxsep}{2.2pt}
    \setlength{\fboxrule}{0.2pt}
    \begin{minipage}{0.24\linewidth}
    \fcolorbox{white}{white}{\includegraphics[width=0.95\linewidth]{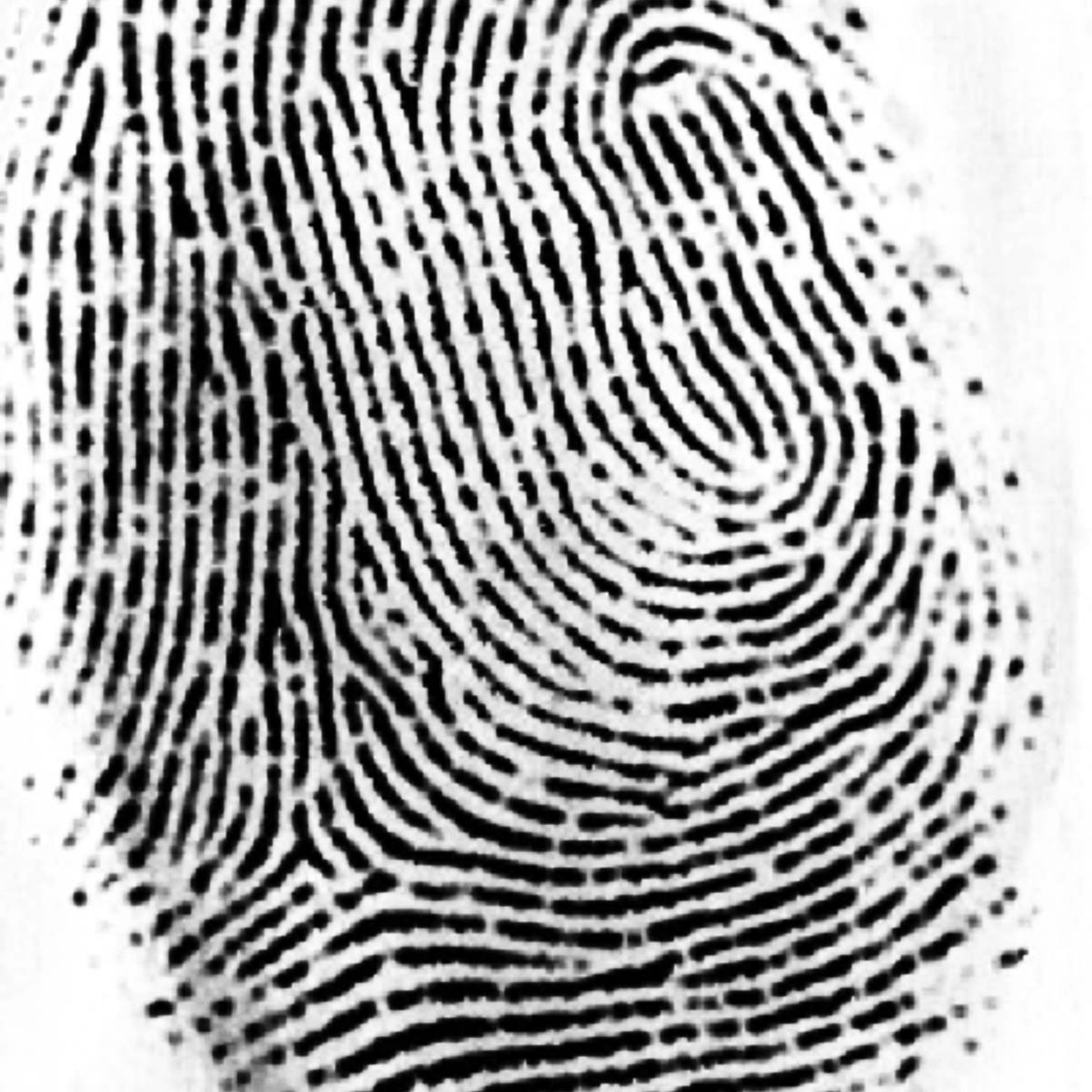}}\\
    \centering {\small(a)}\vspace{0.5em}
    \fcolorbox{white}{white}{\includegraphics[width=0.95\linewidth]{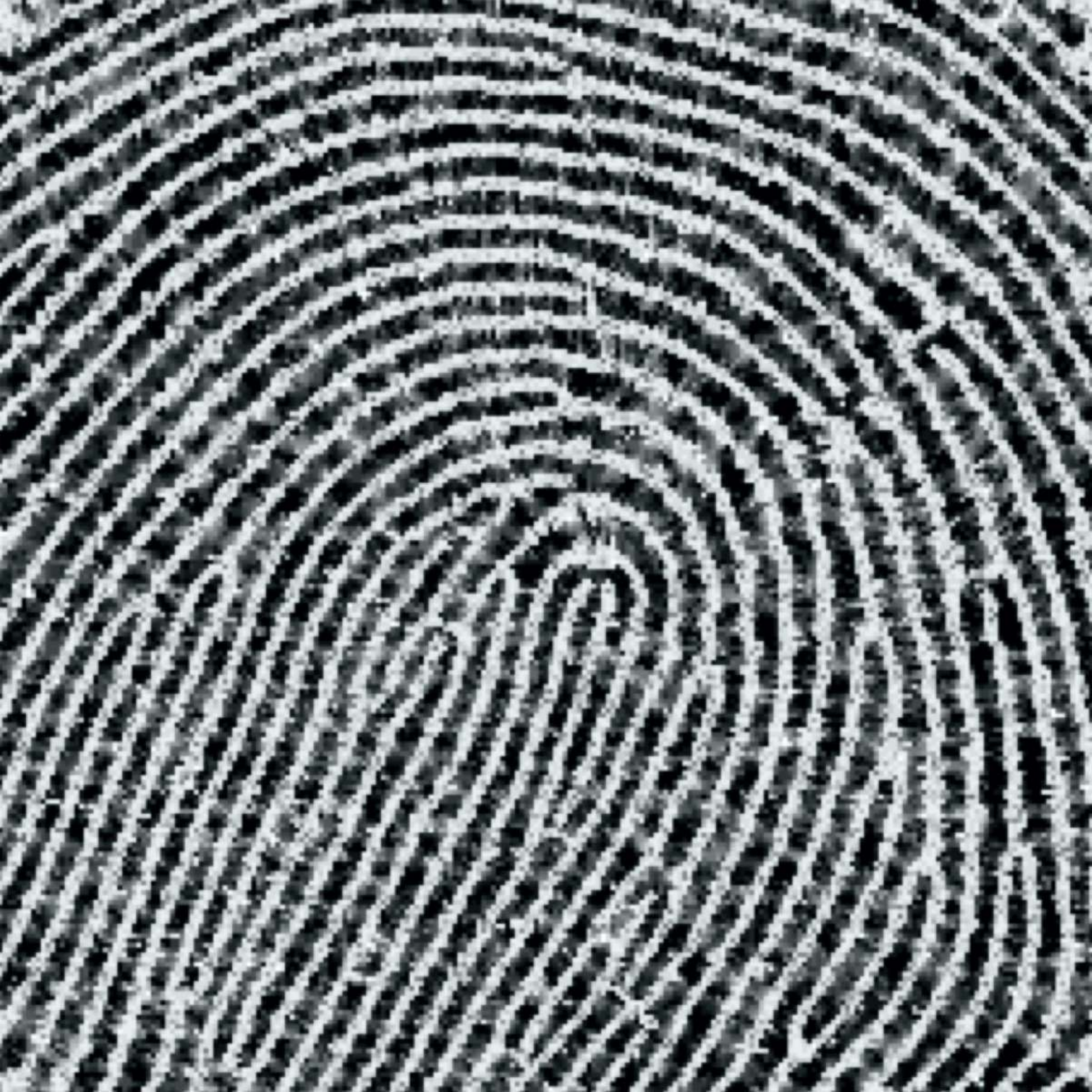}}\\
    \centering {\small (e)}
    \end{minipage}\;
    \begin{minipage}{0.24\linewidth}
    \fcolorbox{white}{white}{\includegraphics[width=0.95\linewidth]{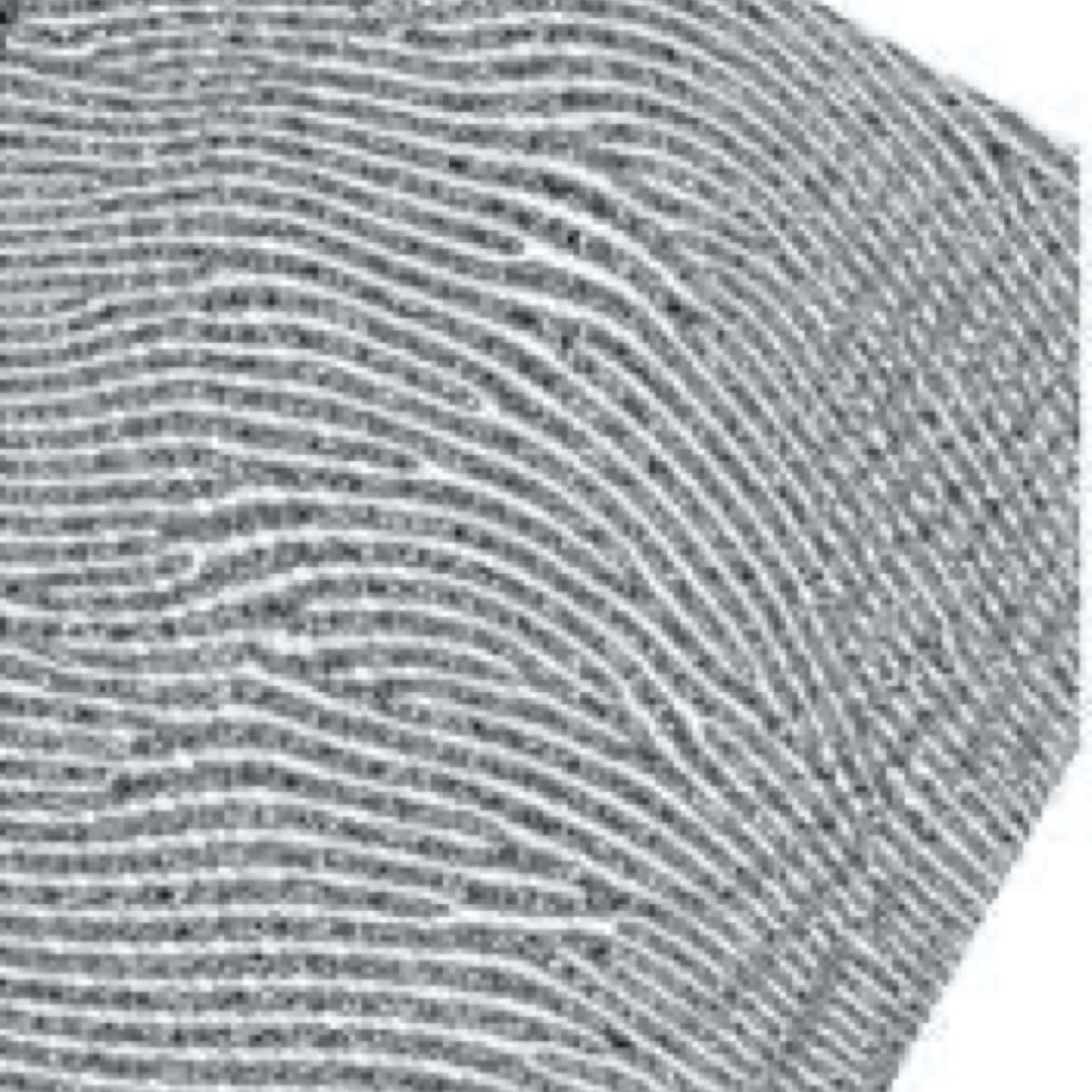}}%
    \\
    \centering {\small(b)}\vspace{0.5em}
    \fcolorbox{white}{white}{\includegraphics[width=0.95\linewidth]{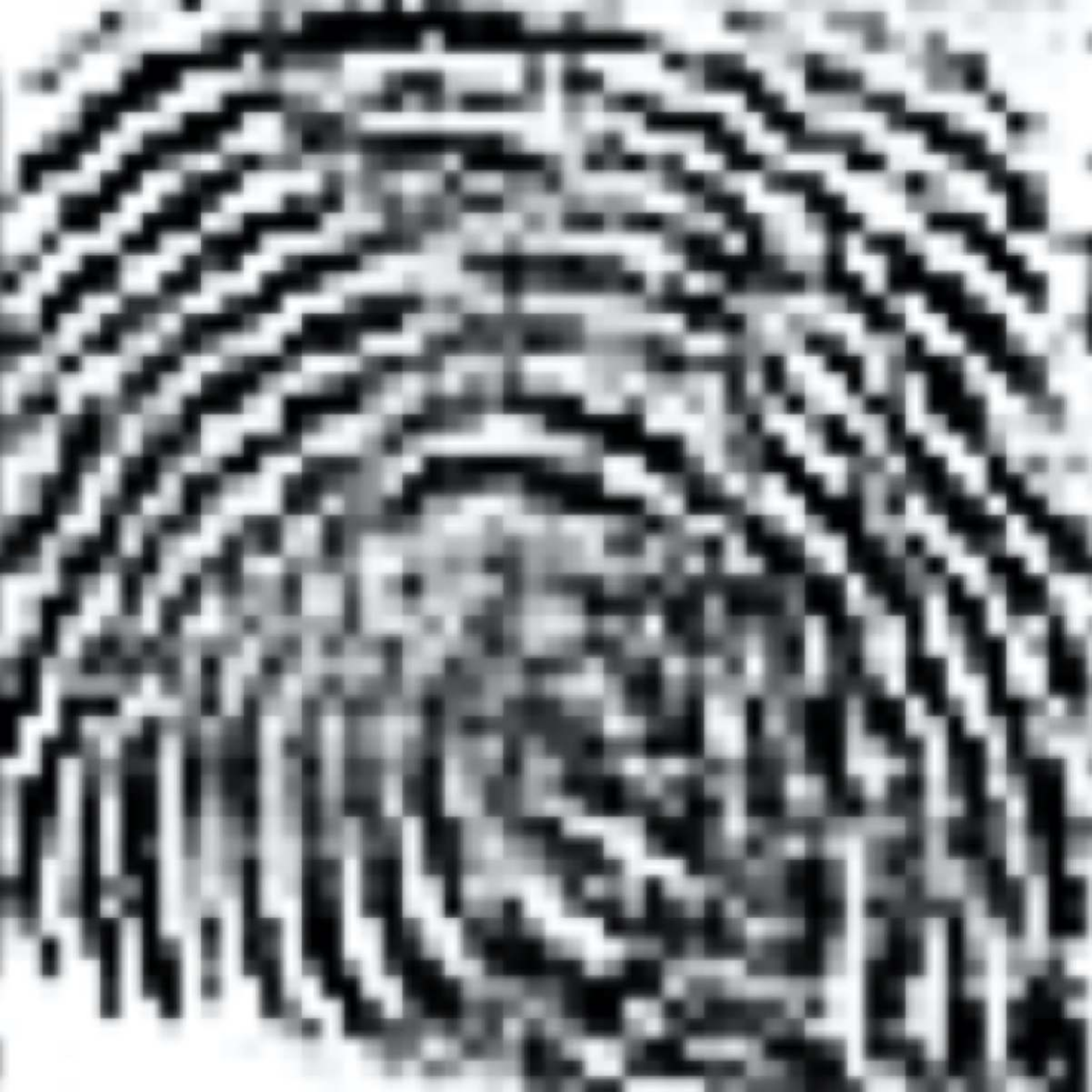}}\\
    \centering {\small(f)}
    \end{minipage}\;
    \begin{minipage}{0.24\linewidth}
    \fcolorbox{white}{white}{\includegraphics[width=0.95\linewidth]{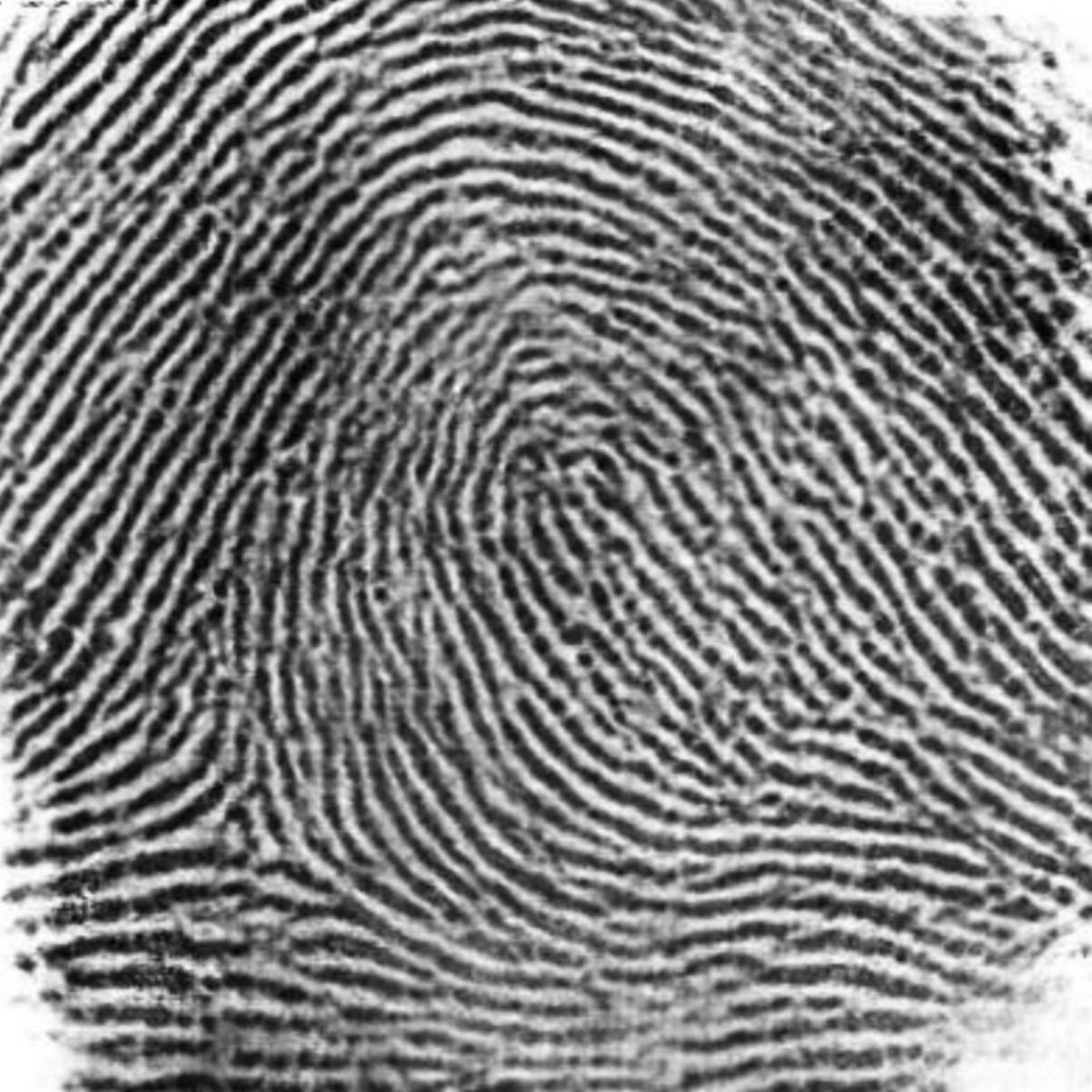}}\\
     \centering {\small(c)}\vspace{0.5em}
    \fcolorbox{white}{white}{\includegraphics[width=0.95\linewidth]{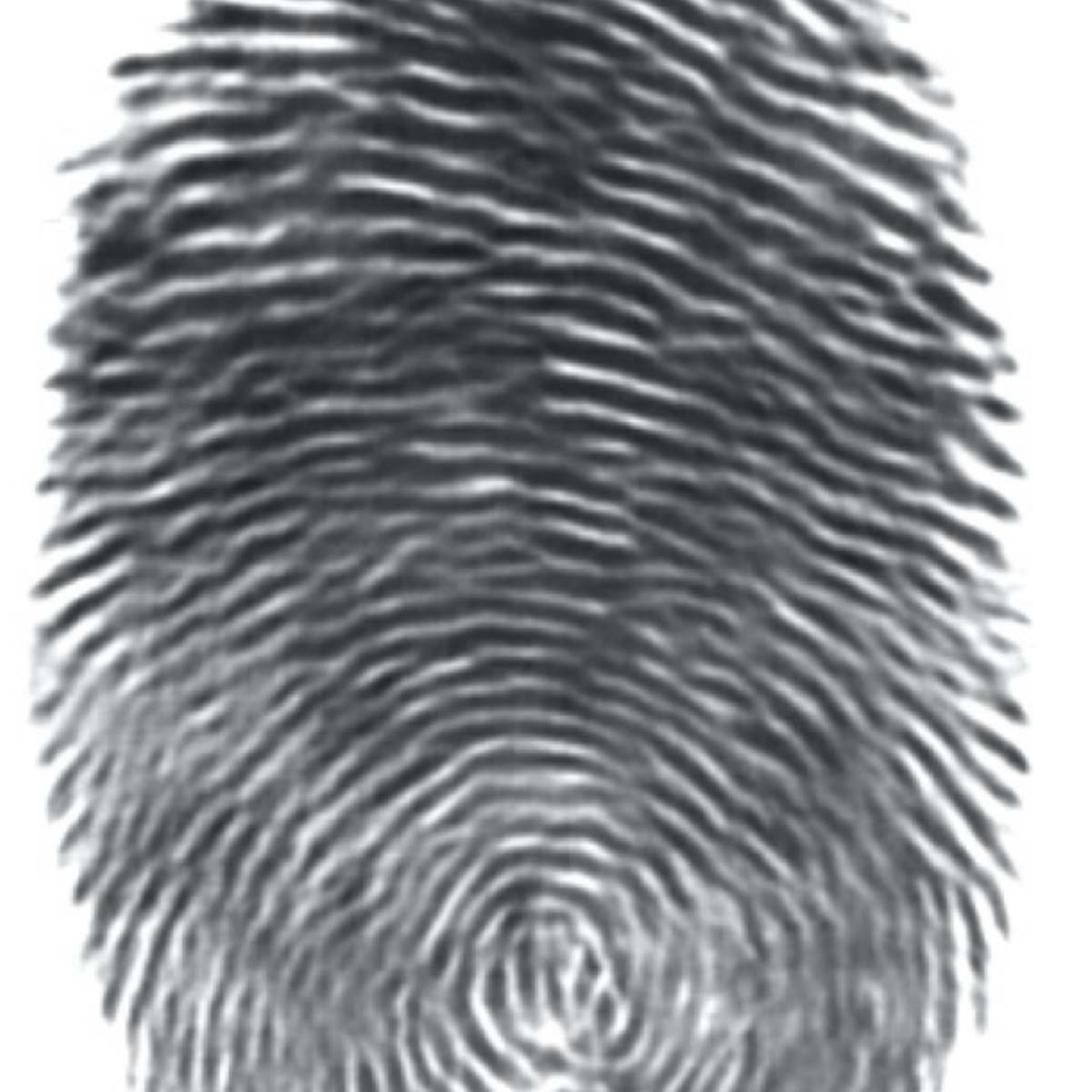}}\\
    \centering {\small(g)}
    \end{minipage}\;
    \begin{minipage}{0.24\linewidth}
    \fcolorbox{white}{white}{\includegraphics[width=0.95\linewidth]{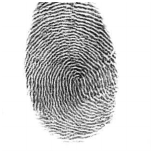}}\\
    \centering {\small(d)}\vspace{0.5em}
    \fcolorbox{white}{white}{\includegraphics[width=0.95\linewidth]{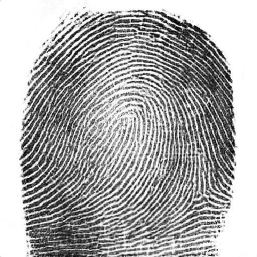}}\\
    \centering {\small(h)}
    \end{minipage}\;
    \captionof{figure}{Comparison of synthetic fingerprints. Fingerprints are synthesized with methods in (a)~\cite{cappeli}, (b)~\cite{zhao}, (c)~\cite{kai_10mil}, (e)~\cite{johnson}, (f)~\cite{fing_gan}, and (g)~\cite{attia}. The plain-print (d) and rolled-print (h) were synthesized using the proposed approach. Qualitatively speaking, our synthetic fingerprints demonstrate more realism than existing approaches. This is further supported based on our quantitative evaluation.}
    \label{fig:fing_examples}
\end{center}
}]
    \item Large-scale fingerprint search evaluation (accuracy and efficiency) against the 100 million synthetic-prints gallery. To the best of our knowledge, this is the largest-scale fingerprint search results ever reported in the literature. 
\end{itemize}

\begin{figure*}
  \includegraphics[width=\linewidth]{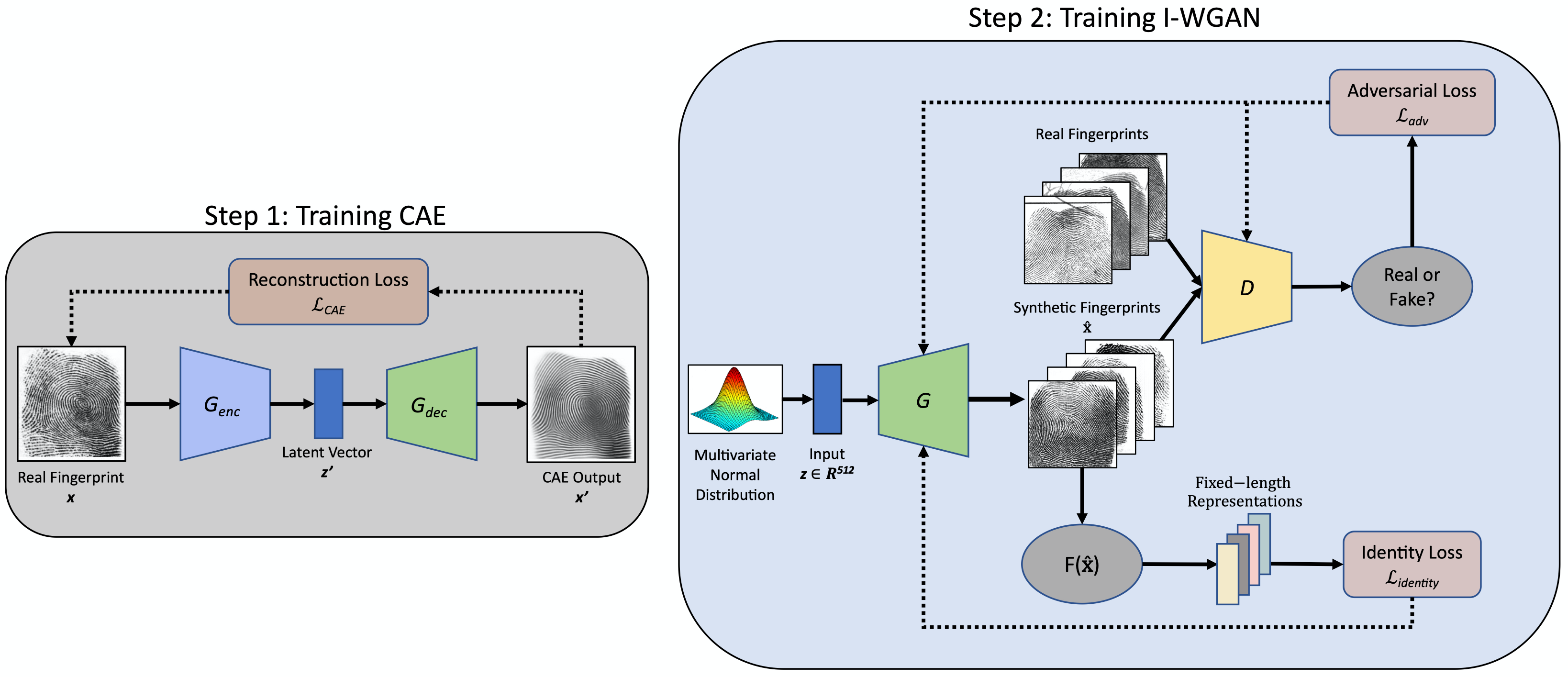}
  \caption{Schematic of the proposed fingerprint synthesis approach. Step 1 shows the training flow of the CAE, which includes the encoder $G_{enc}$ and the decoder $G_{dec}$. Step 2 illustrates the training of the I-WGAN ($G$ and $D$) along with the introduced identity loss. The weights of $G$ were initialized using the trained weights of $G_{dec}$. The solid black arrows show the forward pass of the networks while the dotted black lines show the propagation of the losses.}
  \label{fig:architecture}
\end{figure*}

\section{Related Work}

A plethora of approaches have been proposed for synthetic fingerprint generation which are concisely summarized in Table~\ref{tab:synthesis_literature}. In addition, Figure \ref{fig:fing_examples} shows a qualitative comparison between the synthetic fingerprints generated using~\cite{attia, fing_gan, zhao, cappeli, kai_10mil,johnson} and our proposed approach. These approaches can be broadly categorized into either \textbf{\textit{model-based}} approaches or \textbf{\textit{learning-based}} approaches. While the learning-based methods require large amounts of training data, they are capable of generating more realistic fingerprints than the model-based approaches (shown in our experiments). The limitations of published model-based approaches are further elaborated upon below:

\begin{itemize}[]
    \item Approaches in~\cite{cappeli, johnson, zhao, sfinge, cappelli_iet} assume an independent ridge orientation and minutiae model. As such, minutiae distributions could be generated without having a valid ridge orientation field.
    \item Approaches in~\cite{cappeli, johnson, sfinge, cappelli_iet} assume a fixed fingerprint ridge-width. However, real fingerprints have varying ridge widths. In fact, \cite{svm_fing} reported a 98\% accuracy in discriminating real fingerprints from synthetic fingerprints using only two ridge spacing features.
    \item The ridge structure model in \cite{cappeli, johnson, zhao, sfinge, cappelli_iet} uses a masterprint. Due to Gabor filtering and the AM/FM models, the generated masterprints have non-consistent ridge flow patterns which lead to unrealistic fingerprint images.
    \item The approaches in~\cite{cappeli, johnson, zhao, sfinge, cappelli_iet} are susceptible to generating unrealistic minutiae configurations as their minutiae models do not consider local minutiae configurations. In particular, authors in~\cite{Gottschlich} showed that they could perfectly discriminate between real and synthetic \cite{cappeli} fingerprints using minutiae configurations alone.
    \item None of the approaches generated a fingerprint dataset of size comparable to large-scale operational datasets.
\end{itemize}

More recent learning-based approaches to fingerprint synthesis~\cite{kai_10mil,attia,fing_gan,finger_gan} have utilized generative adversarial networks (GANs)~\cite{gan}. While several of these approaches are capable of improving the realism of fingerprint images using large training datasets of real fingerprints, they do not consider the uniqueness or individuality of the generated fingerprints. Indeed, without any additional supervision, a generator may continually synthesize a small subset of fingerprints, corresponding to only a few identities. This was quantitatively demonstrated in~\cite{kai_10mil} where the search accuracy against real fingerprint galleries was lower than that of synthetic fingerprint galleries. In this work, we build upon the state-of-the-art learning-based synthesis algorithms to exploit on the realism they offer, however, we also incorporate additional supervision (identity loss) to encourage the generation of fingerprints which are of distinct identities. After improving the uniqueness and realism of synthetic fingerprints, we synthesize 100 million fingerprints (the largest gallery reported in the literature) to conduct our large-scale fingerprint search evaluation.

\section{Proposed Approach}

Improved WGAN \cite{iwgan} (I-WGAN) is the backbone of the proposed fingerprint synthesis algorithm. As was done in~\cite{kai_10mil}, we initialize the weights of the I-WGAN generator using the trained weights of the decoder of a convolutional autoencoder (CAE). Our main contribution to the synthesis process is the introduction of an identity loss which guides the generator to synthesize fingerprint images with more distinct characteristics during training.

\begin{figure}[t]
    \begin{subfigure}{.48\linewidth}
        \centering
        \includegraphics[width=\linewidth]{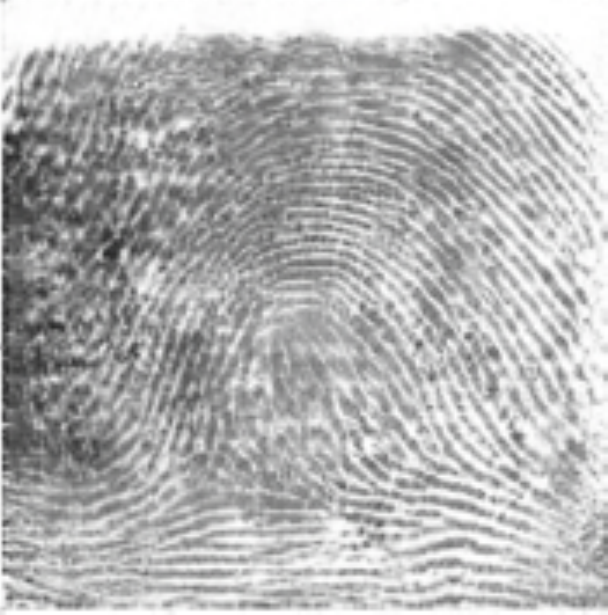}
        \caption{}
        \label{fig:iwgan}
    \end{subfigure}
    \begin{subfigure}{.48\linewidth}
        \centering
        \includegraphics[width=\linewidth]{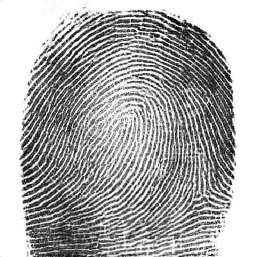}
        \caption{}
        \label{fig:iwgan_init}
    \end{subfigure}
    \caption{Fingerprint image in (a) was generated by $G$ without CAE weight initialization and fingerprints image in (b) was generated by $G$ with CAE weight initialization from $G_{dec}$.}
    \label{fig:iwgan_init_figure}
\end{figure}

The schematic of the proposed approach is shown in Figure \ref{fig:architecture}. The following sub-sections explain the major components of the algorithm in detail.

\subsection{Convolutional Autoencoder}

The first step of our proposed approach involves training a convolutional autoencoder (CAE) \cite{cae} in an unsupervised fashion. The encoder ($G_{enc}$) takes a $512\times512$ grayscale input image ($x$) and transforms it into a 512-dimensional latent vector ($z$'). The decoder ($G_{dec}$) takes the latent vector ($z$') as input and attempts to reconstruct back the input image ($x$'). More formally, the CAE minimizes the reconstruction loss shown in Equation \ref{eq:cae_loss}.
\begin{equation}
    \mathcal{L}_{CAE} = || x - x' ||_{2}^{2} \label{eq:cae_loss}
\end{equation}

After training the CAE, we use the weights of $G_{dec}$ to initialize the Generator $G$ of the I-WGAN. This initialization procedure injects domain knowledge into $G$ enabling it to produce fingerprint images which are of higher quality (as was shown in~\cite{kai_10mil}). Figure \ref{fig:iwgan_init_figure} shows a comparison between fingerprint images generated using $G$ with and without weight initialization by $G_{dec}$.

\subsection{Improved-WGAN}

The architecture of I-WGAN \cite{iwgan} used in our approach follows the implementation of~\cite{dcgan}. Both the generator $G$ and the discriminator $D$ consist of seven convolutional layers with a kernel size of $4\times4$ and a stride of $2\times2$ (architecture deferred to supplementary). $G$ takes a 512-dimensional latent vector $z$$\sim$$\mathbb P_{z}$ as input, where $\mathbb P_{z}$ is a multivariate Gaussian distribution, and outputs a $512\times512$ dimensional grayscale image. All the convolutional layers in $G$ have ReLU as their activation function, except the last layer which uses \textit{tanh}. The discriminator $D$ has the inverse architecture of the generator $G$. All the convolutional layers in $D$ use LeakyReLU as the activation function. Moreover, batch normalization is applied to all the layers of $G$ and $D$.

\subsection{Identity Loss}
The authors in~\cite{kai_10mil} demonstrated that synthetic fingerprints are less unique than real fingerprints by showing that the search accuracy on NIST SD4 \cite{nist_sd4} (2,000 rolled fingerprint pairs) against 250K synthesized rolled fingerprints is higher than when 250K real rolled fingerprints are used as the gallery. One plausible explanation for this difference in search accuracies is that no distinctiveness measure was used during training of the synthetic fingerprint generator, leading to lower ``diversity” among the synthesized fingerprints than among real fingerprints. To rectify this, we introduce an identity loss to encourage synthesis of more distinct or unique fingerprint images.

The authors in~\cite{deepprint} introduced a deep network, called DeepPrint, to extract a fixed-length representation from a fingerprint image, encoding the minutiae as well as the texture information (i.e. the identity) of the fingerprint into a 192-dimensional vector. Since the DeepPrint feature extractor is a differentiable function (i.e. it is a deep network), we can use the trained DeepPrint~\cite{deepprint} network ($F(x)$) during the training process (computing the loss) of the I-WGAN (unlike DeepPrint, COTS minutiae matchers are not differentiable and therefore cannot be used to train the network). In particular, for each mini-batch, we use DeepPrint to extract the fixed-length representations of all the fingerprint images ($x$) synthesized by the generator $G$. The similarity between each pair of representations in the mini-batch is then minimized in order to guide $G$ to produce fingerprints with different identities, i.e. more distinct fingerprints.

More formally, for each pair of latent vectors $(z_i,z_j)$ in a mini-batch, the identity loss, which is given by Equation. \ref{eq:identity_loss}, is minimized.
\begin{equation}
    \mathcal{L}_{identity} = \frac{1}{\sum \left \| F(G(z_i)) - F(G(z_j)) \right \|_{2}},\left ( z_i \neq z_j \right) \label{eq:identity_loss}
\end{equation}

\subsection{Training Details}

The CAE and the I-WGAN were both trained using 280,000 rolled fingerprint images from a law enforcement agency~\cite{kai_fing}. The fingerprints in the training set range in size from $407\times346$ pixels to $750\times800$ pixels at 500 dpi. We coarsely segment them to $512\times512$ pixels since an area of $512\times512$ pixels with a resolution of 500 dpi is sufficient to cover the entire friction ridge pattern. We acquired the weights for DeepPrint from the authors in~\cite{deepprint}.

\begin{figure*}
  \includegraphics[width=\linewidth]{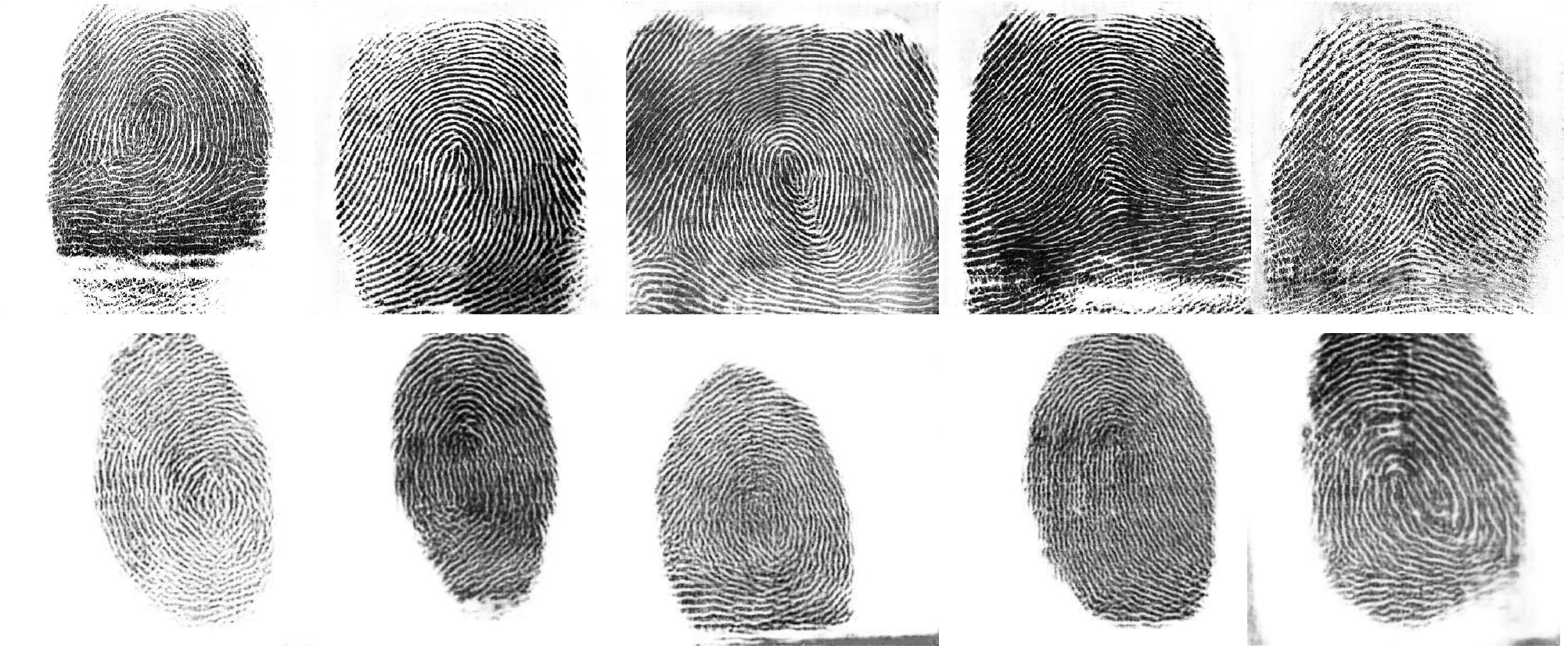}
  \caption{Examples of fingerprint images ($512\times512$ @ 500 dpi) synthesized by our approach. The top row corresponds to synthetic rolled prints while the bottom row shows synthetic plain fingerprints.}
  \label{fig:collage}
\end{figure*}

\subsection{Fine-Tuning for Plain-Prints}

The CAE and I-WGAN are trained to synthesize rolled fingerprints. However, we show that by fine-tuning I-WGAN on an aggregated dataset of 84K plain-prints from~\cite{gct, contactless}, we can also generate high-quality plain-prints. This enables us to compare the realism of our learning-based synthetic fingerprints more fairly with existing, publicly available plain-print synthesis algorithms (SFinGe~\cite{sfinge}).

\subsection{Synthesis Process}

After training I-WGAN with the identity loss, we generate 100 million rolled fingerprint images. Generating 100 million fingerprint images is non-trivial and requires significant memory and processor resources. Our synthesis was expedited via a High Performance Computing Center (HPCC) on our campus. In particular, we ran 100 jobs in parallel on the HPCC server, with each job generating 1 million fingerprint images. As each fingerprint was generated, we simultaneously extracted its DeepPrint~\cite{deepprint} fixed-length representation (to be used in our later search experiments). Each job was assigned an Intel Xeon E5-2680v4@2.40GHz CPU, 32GB of RAM, and a {NVIDIA} Tesla K80 GPU. This took 51 CPU hours\footnote{Since the HPCC follows a queue-based approach for running jobs, all the 100 jobs were initiated within 4 days.}. Finally, we also generated 2,000 plain-prints using our fine-tuned plain-print model for comparison with other synthetic plain-print approaches.


\section{Experimental Results}

In our experimental results, we first demonstrate that our synthetic fingerprints are more realistic and unique than those from published methods. Then, we show the fingerprint search results of COTS and DeepPrint~\cite{deepprint} matchers against a background of 1 million and 100 million synthetic fingerprints respectively.

\subsection{Fingerprint Realism}

\begin{table*}[t]
\small
    \centering
    \begin{tabular}{ P{1.5cm}P{7.0cm}P{7.0cm} }
    \toprule
    {} & \textbf{Plain Fingerprint datasets} & \textbf{Rolled Fingerprint datasets} \\
    \toprule
    \textbf{Real} & 
    \begin{itemize}[nolistsep, before*={\mbox{}\vspace{-\baselineskip}}, after*={\mbox{}\vspace{-\baselineskip}}]
        \item CASIA-Fingerprint V5 \cite{casia} (2000 fingerprints)
        \item NIST SD302L \cite{nist_sd302} (1951 fingerprints)
        \item NIST SD302M \cite{nist_sd302} (1979 fingerprints)
    \end{itemize}{} & 
    \begin{itemize}[nolistsep, before*={\mbox{}\vspace{-\baselineskip}}, after*={\mbox{}\vspace{-\baselineskip}}]
        \item NIST SD4 \cite{nist_sd4} (2000 enrollment fingerprints)
        \item NIST SD14 \cite{nist_sd14} (last 2000 enrollment fingerprints)
        \item NIST SD302U \cite{nist_sd302} (2000 fingerprints)
    \end{itemize}{} \\
    \midrule
    \textbf{Synthetic} & 
    \begin{itemize}[nolistsep, before*={\mbox{}\vspace{-\baselineskip}}, after*={\mbox{}\vspace{-\baselineskip}}]
        \item SFinGe \cite{sfinge} (2000 fingerprints)
        \item Proposed Approach (2000 fingerprints)
    \end{itemize}{} & 
    \begin{itemize}[nolistsep, before*={\mbox{}\vspace{-\baselineskip}}, after*={\mbox{}\vspace{-\baselineskip}}]
        \item Cao and Jain~\cite{kai_10mil} (2000 fingerprints)
        \item Proposed Approach (2000 fingerprints)
    \end{itemize}{} \\
    \bottomrule
    \end{tabular}
    \caption{Real and synthetic fingerprint datasets that were used for comparison in our experiments.}
    \label{tab:datasets_plain_rolled}
\end{table*}

To evaluate the similarity of synthetic fingerprints to real fingerprints, we used the fingerprint datasets enumerated in Table \ref{tab:datasets_plain_rolled}. In particular, we compare real plain-prints to synthetic plain-prints and real rolled-prints to synthetic rolled-prints.

\noindent~\textbf{\textit{Metrics: }}~Four template-level and three block-level comparison indicators used for this evaluation are taken from~\cite{cappelli_iet}, including, minutiae count - template and block, direction - template and block, convex hull area, spatial distributions (minutiae spatial distributions represented as a 2D minutiae histogram~\cite{Gottschlich, kai_10mil}), and block minutiae quality. In addition, we utilize the NFIQ 2.0~\cite{nfiq} quality scores distribution. Minutiae points (necessary to compute these metrics) were extracted using state-of-the-art COTS SDKs: VeriFinger v11.0 \cite{verifinger} and Innovatrics v.7.6.0.627 \cite{innovatrics}.

\noindent~\textbf{\textit{Statistical Test: }}~We use the Kolmogorv-Smirnov test~\cite{ks_test} to compute the difference between the distributions of each of the aforementioned metrics extracted from a set of real fingerprints (benchmark distribution), and synthetic fingerprints (comparison distribution). A low test statistic value indicates high similarity between real fingerprints and the synthetic fingerprints. We compared each synthetic fingerprint dataset with three real fingerprint datasets on all the above-mentioned metrics and calculated the average KS test statistic from the three comparisons. 

\begin{figure*}[t]
    \begin{subfigure}{.49\textwidth}
        \centering
        \includegraphics[width=0.97\linewidth]{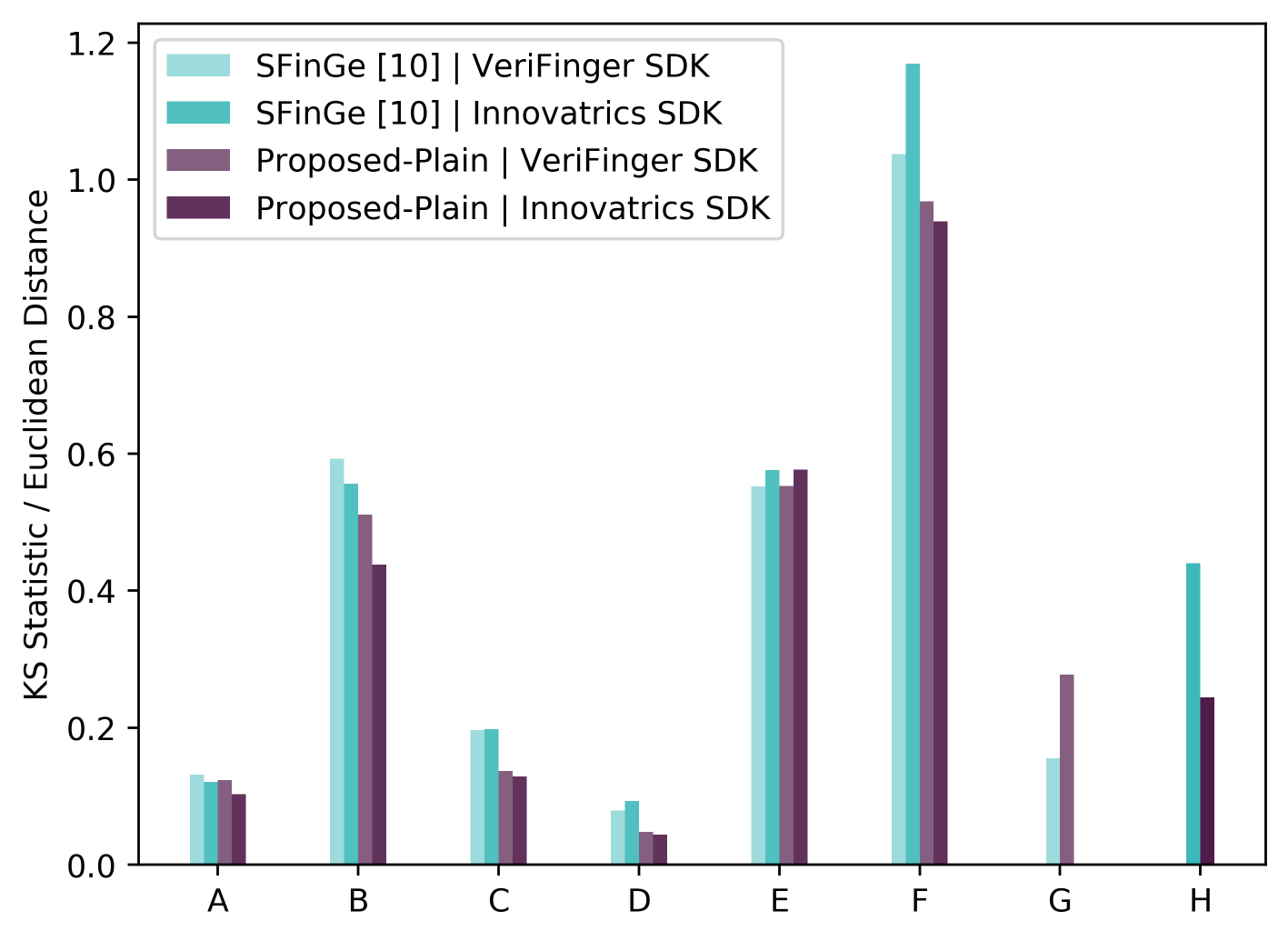}
        \label{fig:compare_plain}
        \caption{}
    \end{subfigure}
    \begin{subfigure}{.49\textwidth}
        \centering
        \includegraphics[width=0.97\linewidth]{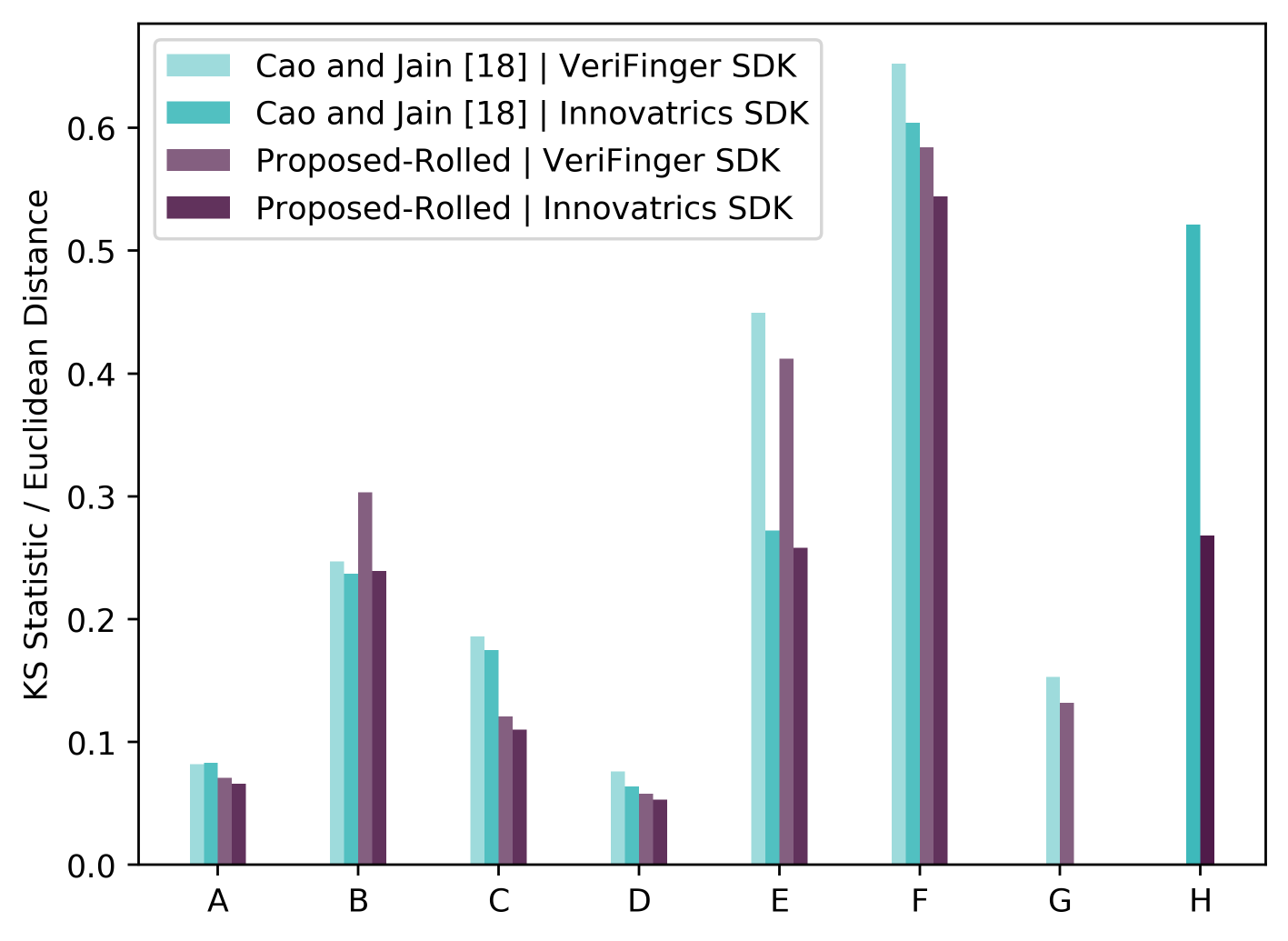}
        \label{fig:compare_rolled}
        \caption{}
    \end{subfigure}
    \caption{The comparison of synthetic plain (a) and rolled fingerprints (b) to real fingerprints using 8 metrics (minutiae count - block [A] and template [B], direction - block [C] and template [D], convex hull area [E], spatial distributions [F], block minutiae quality [G], and NFIQ 2.0 [H]) and a Kolmogorv-Smirnov test. A lower value indicates higher similarity of synthetic fingerprints to real fingerprints.}
    \label{fig:compare}
\end{figure*}

\noindent~\textbf{\textit{Realism Results: }}Figure~\ref{fig:compare} shows the KS test statistics between our synthetic rolled and plain prints and real rolled and plain prints as well as the comparison with state-of-the-art methods for plain-print synthesis - SFinGe v5.0~\cite{sfinge} (21.1mm$\times$28.4mm acquisition area, 500 DPI resolution, 416$\times$560 image dimensions, natural class distribution), and rolled-print synthesis~\cite{kai_fing}. We note that our synthetic plain-prints are more similar to real plain-prints on 7 out of the total 8 indicators than state-of-the-art synthetic plain-prints~\cite{sfinge}. Similarly, our synthetic rolled-prints are more similar to real rolled-prints than state-of-the-art synthetic rolled-prints~\cite{kai_10mil} on 7 out of the 8 indicators.

\subsection{Fingerprint Uniqueness}

\subsubsection{Imposter Scores Distribution}

\begin{figure}[t]
  \includegraphics[width=\linewidth]{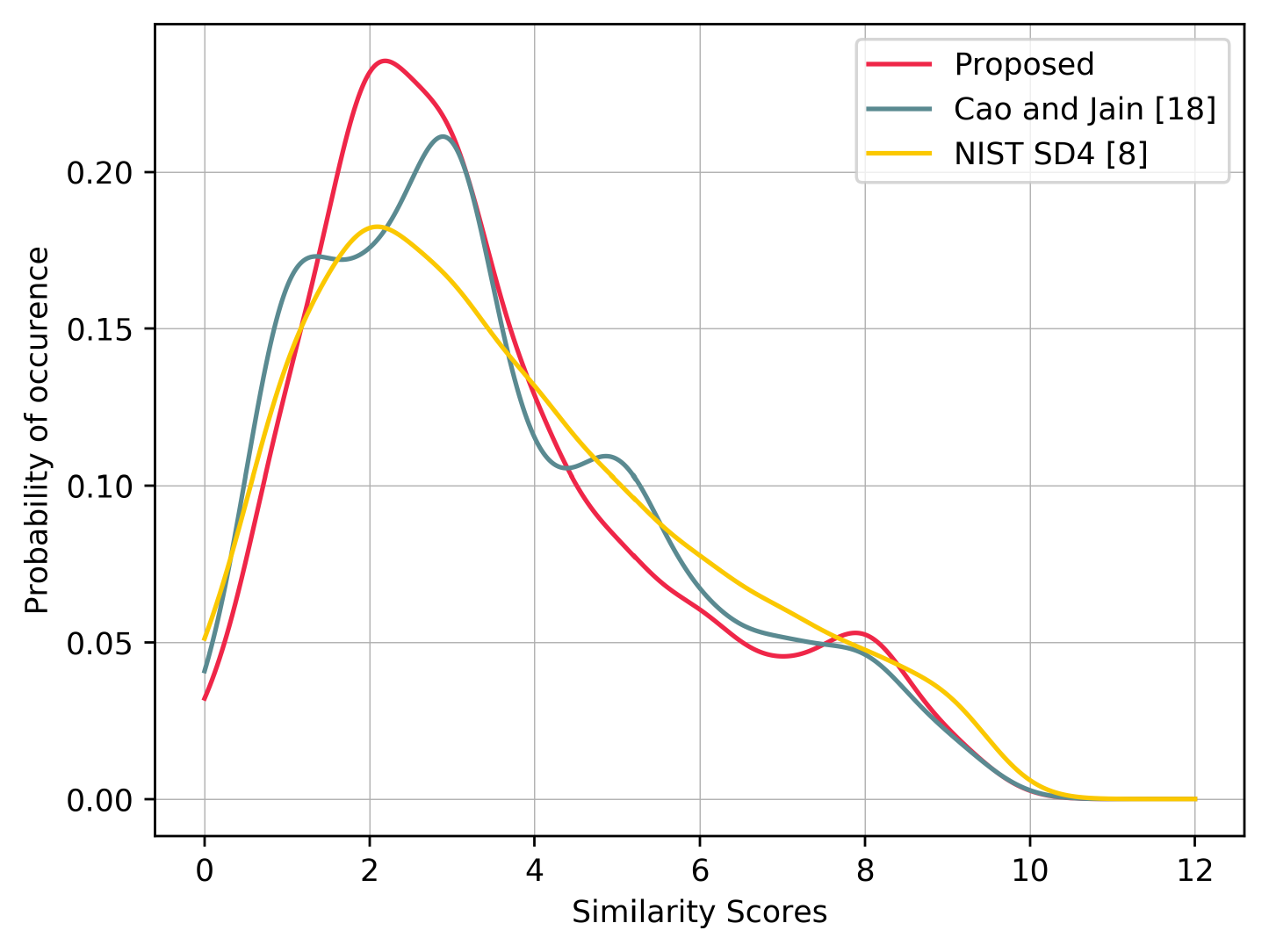}
  \caption{Imposter score distributions computed using the proposed synthetic rolled-prints, synthetic rolled-prints from~\cite{kai_10mil} and real fingerprints from NIST SD4~\cite{nist_sd4}. Scores were computed using VeriFinger SDK v11.}
  \label{fig:imposter_scores}
\end{figure}

To determine the diversity of our synthetic fingerprints (in terms of identity), we first computed 500K imposter scores using Verifinger in conjunction with our synthetic rolled-prints and also the synthetic rolled-prints from~\cite{kai_10mil} (Figure~\ref{fig:imposter_scores}). We note that the mean (3.47) and standard deviation (2.13) of the imposter scores computed when using our synthetic rolled-prints are both lower than the mean (3.48) and standard deviation of (2.18) of the imposter scores computed with synthetic rolled-prints from~\cite{kai_10mil}. We also note a higher peak in the imposter distribution from our synthetic rolled-prints at lower similarity values (Figure~\ref{fig:imposter_scores}). This indicates that our addition of an identity loss has helped guide our synthetic rolled-prints to be more unique.

\subsubsection{DeepPrint Search Against 1 Million}

\begin{figure}[t]
  \includegraphics[width=\linewidth]{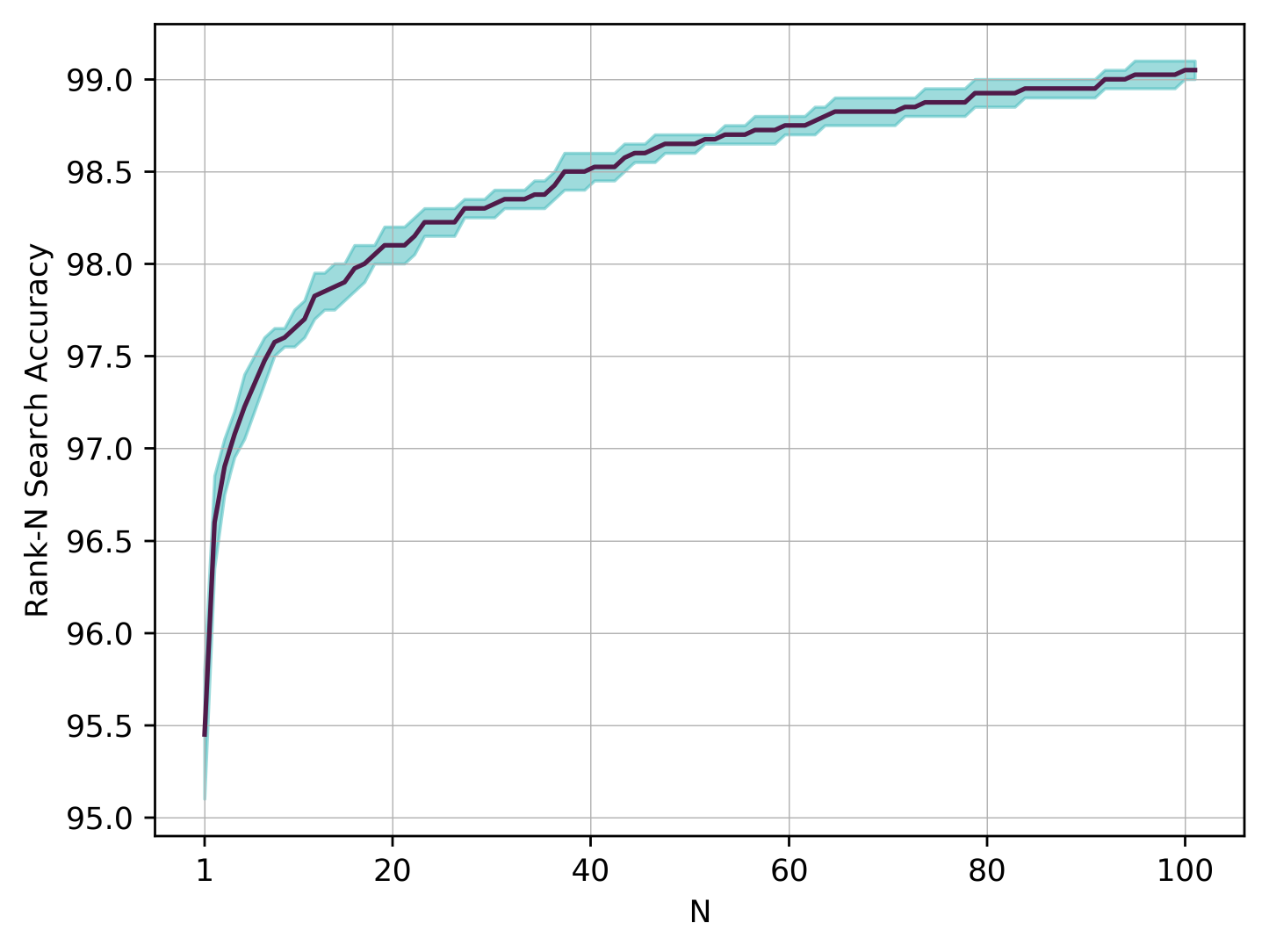}
  \caption{Confidence interval for rank-N search accuracy on NIST SD4 \cite{nist_sd4}. 10-folds of 1 million synthetic rolled fingerprints were used to compute confidence intervals.}
  \label{fig:conf_interval}
\end{figure}

We also demonstrate uniqueness by conducting large-scale search experiments. More uniqueness in the gallery leads to a lower search accuracy. First, we obtain an ``upper-bound" on search performance using NIST SD4 \cite{nist_sd4} in conjunction with 10 different subsets of 1 million synthetic rolled-prints from our proposed approach. We use DeepPrint~\cite{deepprint} as the matcher. Confidence intervals for rank-N search accuracies are shown in Figure \ref{fig:conf_interval}. The mean rank-1 search accuracy is $95.53\%$ with confidence interval of $[95.1, 95.8]$.

\subsubsection{COTS Search Against 1 Million}

We also evaluate the distinctiveness of our synthetic fingerprints using the Innovatrics SDK. In particular, we perform fingerprint search using NIST SD4 \cite{nist_sd4} in conjunction with our synthetic rolled-prints and synthetic rolled-prints from~\cite{kai_10mil}. The rank-1 search accuracies for both the datasets (at gallery sizes of 250K and 1 Million) are shown in Table \ref{tab:inn_rank_1}. From Table~\ref{tab:inn_rank_1} we note that although the search accuracy is lower when using rolled-prints from~\cite{kai_fing} at a lower gallery size of 250K gallery, the search performance drops more rapidly as the synthetic gallery scales to 1 Million when using our proposed synthetic rolled-prints. This suggests that the uniqueness of our synthetic rolled-prints becomes more evident at large gallery sizes~\cite{kai_fing}. We also acknowledge that there is still room for further improvement in the ``uniqueness" of synthetic fingerprints as the rank-1 search performance of Innovatrics against 1.1 Million real rolled-prints was reported to be 89.2\% in~\cite{deepprint} (vs. 90.35\% against 1 Million of our synthetic rolled-prints).

\begin{table}[t]
\small
    \centering
    \begin{tabular}{ P{1.9cm}P{2.8cm}P{2.8cm} }
    \toprule
    {} & \textbf{Proposed Approach} & \textbf{Cao \& Jain~\cite{kai_10mil}} \\
    \toprule
    \textbf{250K Gallery} & 91.45\% & 90.85\% \\
    \midrule
    \textbf{1M Gallery} & 90.35\% & 90.40\% \\
    \bottomrule
    \end{tabular}
    \caption{Innovatrics rank-1 fingerprint search accuracies on NIST SD4~\cite{nist_sd4} using synthetic rolled-prints as gallery.}
    \label{tab:inn_rank_1}
\end{table}

\subsubsection{Search Against 100 Million}

To date, no work has been done in the literature to evaluate fingerprint search performance at a scale of 100 Million. Two main challenges preventing this important evaluation are (i) the computational challenges in generating 100 million fingerprints and (ii) finding a matcher fast enough to search 100 million. Our use of a high-performance computing center solves the former, and the recently proposed fast fixed-length matcher, DeepPrint \cite{deepprint} addresses the latter. Therefore, in this paper, we not only synthesize fingerprints (like others in the literature), but we also scale the synthesis to 100 million \textbf{and} we actually show the search performance against the large-scale synthesized gallery.

\begin{figure}[t]
  \includegraphics[width=\linewidth]{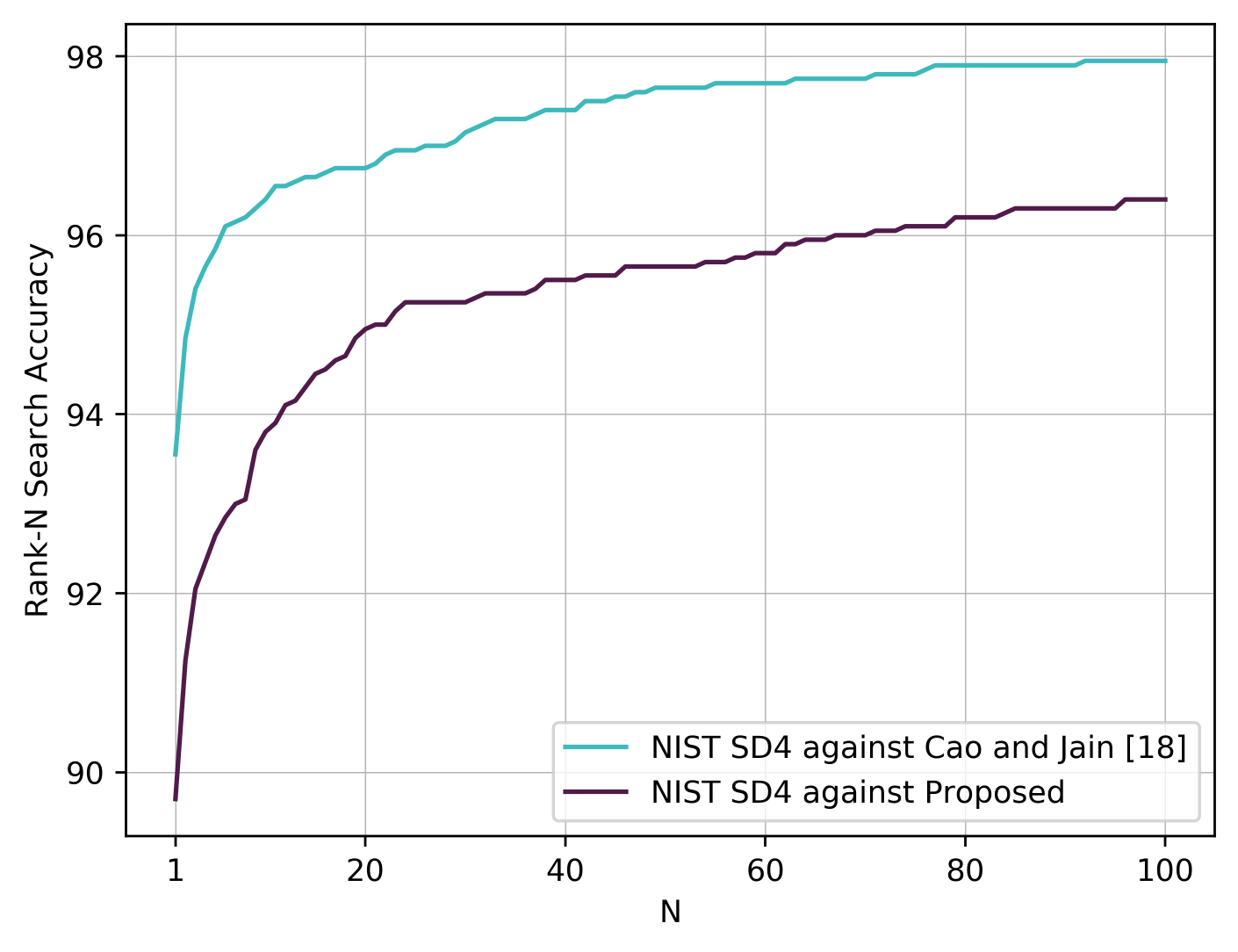}
  \caption{Rank-N fingerprint search accuracies on NIST SD4~\cite{nist_sd4} for the augmented gallery of 100 million fingerprints using DeepPrint \cite{deepprint} as the fingerprint matcher.}
  \label{fig:dp_rank_n}
\end{figure}

Figure~\ref{fig:dp_rank_n} shows the rank-N search accuracies on NIST SD4~\cite{nist_sd4} in conjunction with 100 million of our synthetic rolled-prints (rank-1 accuracy of 89.7\%). We also show the search performance when using 100 million synthetic rolled-prints from~\cite{kai_fing} as a benchmark (rank-1 accuracy of 93.55\%). We note that at a gallery size of 100 million, the search performance is much lower when using our synthetic fingerprints than when using synthetic fingerprints from~\cite{kai_fing}. This is strong evidence that our fingerprints are more unique than the fingerprints from~\cite{kai_fing}. 

Since DeepPrint performed similarly to top-COTS matchers on a gallery of 1 million real fingerprints in~\cite{deepprint}, we posit that DeepPrint and by extension COTS matchers have room for improvement with galleries of hundreds of millions or even billions of fingerprints (e.g. Aadhaar). This is our ongoing research direction. Note that we could not evaluate COTS against 100 million due to the inadequate speed of the SDK licenses that we have for the matchers (about 400 days for a rank-1 search on a gallery size of 100 million images as compared to 11s for DeepPrint).

\section{Conclusions}

Given the difficulty in obtaining large-scale fingerprint datasets for evaluating large-scale search, we propose a fingerprint synthesis algorithm which utilizes I-WGAN and an identity loss to generate diverse and realistic fingerprint images. We show that our method can be used to synthesize realistic rolled-prints as well as plain-prints. We demonstrate that our synthetic fingerprints are more realistic and diverse than current state-of-the-art synthetic fingerprints. Finally, we show, for the first time in the literature, fingerprint search performance at a scale of 100 million. Our results suggest that state-of-the-art fingerprint matchers must be further improved to operate at a scale of 100 million. Our ongoing work will (i) further scale our search experiments to a gallery of 1 billion, and (ii) further improve the realism and uniqueness of synthesized fingerprints.

{\small
\bibliographystyle{ieee}
\bibliography{submission_example}
}

\clearpage

\appendix
\renewcommand*\appendixpagename{Appendix}
\renewcommand*\appendixtocname{Appendix}
\appendixpage

\section{Improved-WGAN Architecture}

The architecture of I-WGAN \cite{iwgan} used in our approach follows the implementation of~\cite{dcgan}. Both the generator $G$ and the discriminator $D$ consist of seven convolutional layers with a kernel size of $4\times4$ and a stride of $2\times2$ which are detailed in table \ref{tab:iwgan_arch}.

\begin{table}[h]
\small
    \centering
    \begin{tabular}{ p{2.4cm}p{2.5cm}p{1.5cm} }
    \toprule
    \multicolumn{3}{c}{\textbf{Generator $\textbf{\textit{G/G\textsubscript{dec}}}$ architecture}}\\
    \toprule
    \textbf{Operation} & \textbf{Output size} & \textbf{Filter size} \\
    \toprule
    Input & $512\times1$ & \\
    \midrule
    Fully Connected & $16,384 \times 1$ & \\
    \midrule
    Reshape & $4 \times 4 \times 1024$ & \\
    \midrule
    Deconvolution & $8 \times 8 \times 512$ & $4 \times 4$\\
    \midrule
    Deconvolution & $16 \times 16 \times 256$ & $4 \times 4$\\
    \midrule
    Deconvolution & $32 \times 32 \times 128$ & $4 \times 4$\\
    \midrule
    Deconvolution & $64 \times 64 \times 64$ & $4 \times 4$\\
    \midrule
    Deconvolution & $128 \times 128 \times 32$ & $4 \times 4$\\
    \midrule
    Deconvolution & $256 \times 256 \times 16$ & $4 \times 4$\\
    \midrule
    Deconvolution & $512 \times 512 \times 1$ & $4 \times 4$\\
    \bottomrule
    \end{tabular}
    \caption{Architecture of \textit{G/G\textsubscript{dec}} in I-WGAN. The discriminator architecture $D$ is an inverse of the architecture $G$.}
    \label{tab:iwgan_arch}
\end{table}{}

\section{Training Details}

The training of both CAE and I-WGAN were implemented using Tensorflow on an Intel Core i7-8700K @ 3.70GHz CPU with a RTX 2080 Ti GPU. The weights were randomly initialised from a gaussian distribution with mean 0 and standard deviation of 0.02. The optimizer function for the CAE was Adam \cite{adam} with a fixed learning rate of 0.0002, $\beta_1$ as 0.5, and $\beta_2$ as 0.9. The batch size was 128 and the number of training steps were 39,000. The Adam optimizer \cite{adam} was also used in the training of the I-WGAN with a fixed learning rate of 0.0001, $\beta_1$ as 0.0 and $\beta_2$ as 0.9. For I-WGAN, the batch size was set to 32 with 54,000 training steps. However, when I-GAN was finetuned to the dataset of plain fingerprints, it was trained for 37,000 steps.

\end{document}